\documentclass{article}

\usepackage[preprint]{arxiv_v1}

\usepackage[utf8]{inputenc} %
\usepackage[T1]{fontenc}    %
\usepackage{hyperref}       %
\usepackage{url}            %
\usepackage{booktabs}       %
\usepackage{amsfonts}       %
\usepackage{nicefrac}       %
\usepackage{microtype}      %
\usepackage{xcolor}         %
\usepackage{graphicx}
\usepackage{wrapfig}
\usepackage{amsmath}
\usepackage{subcaption}
\usepackage{multirow}

\newcommand{\distlearner}{\textit{Distance Learner}}
\newcommand{\stdclf}{\textit{Standard Classifier}}
\newcommand{\advclf}{\textit{Robust Classifier}}
\newcommand{\expnumber}[2]{{#1}\mathrm{e}{#2}}

\newcommand{\loss}[3]{${#1}~ {\scriptstyle (\pm {#2})} \times 10^{#3}$}

\title{Distance Learner: Incorporating Manifold Prior to Model Training}

\author{%
  Aditya Chetan \\
  Microsoft Research India\\
   \And
   Nipun Kwatra \\
   Microsoft Research India \\
}

\begin{document}

\maketitle

\begin{abstract}

The manifold hypothesis (real world data concentrates near low-dimensional manifolds) is suggested as the principle behind the effectiveness of machine learning algorithms in very high dimensional problems that are common in domains such as vision and speech. Multiple methods have been proposed to explicitly incorporate the manifold hypothesis as a prior in modern Deep Neural Networks (DNNs), with varying success. In this paper, we propose a new method, \distlearner{}, to incorporate this prior for DNN-based classifiers. \distlearner{} is trained to predict the \textit{distance} of a point from the underlying manifold of each class, rather than the class label. For classification, \distlearner{} then chooses the class corresponding to the closest predicted class manifold. \distlearner{} can also identify points as being out of distribution (belonging to neither class), if the distance to the closest manifold is higher than a threshold. We evaluate our method on multiple synthetic datasets and show that \distlearner{} learns much more meaningful classification boundaries compared to a standard classifier. We also evaluate our method on the task of adversarial robustness, and find that it not only outperforms standard classifier by a large margin, but also performs at par with classifiers trained via state-of-the-art adversarial training.

\end{abstract}

\section{Introduction}

Most real world classification problems involve very high dimensional data inputs. However, learning in high dimensions suffers from what is popularly known as the \textit{curse of dimensionality}. Consider, for example, a simple nearest neighbor based-classifier. For a new point to be ``reasonably" close to a point in the training dataset, the number of points in the dataset need to be exponential in the data dimension. This can become prohibitive very quickly. Fortunately, as per the manifold hypothesis, real world data is believed to concentrate near low-dimensional manifolds even though the data is represented in much higher dimensions. ~\citet{pedro_few_useful_things} calls this the ``blessing of non-uniformity" as learners can implicitly or explicitly take advantage of the much lower effective dimension.

Deep Neural Networks (DNNs) have been extremely successful in achieving state-of-the-art performance on high dimensional classification tasks across multiple domains --- e.g, image \citep{cocaimgclf}, text \citep{devlin-etal-2019-bert, 2020t5} and speech \citep{Shen2018NaturalTS}. A remarkable property of DNNs is their ability to generalize well on unseen test data. Understanding the generalization properties of DNNs is an open area. The famous ``no free lunch" theorems~\citep{wolpert1997no_free_lunch_optimization,wolpert1996lack_of_apriori,wolpert2002supervised_no_free_lunch} state that on an average no learner can beat random guessing over all the possible learnt functions. Then, what makes DNNs so successful? Fortunately, the underlying functions for these real world tasks are not sampled randomly from the set of all possible functions. What makes learning work is incorporating some prior knowledge about the underlying function into the learning. Even simple priors such as smoothness can do very well. For example, neural networks encode the smoothness prior via construction of the network where all building blocks are smooth. Convolutional Neural Networks (CNN) encode the prior knowledge that the underlying classification function should be translation invariant. Similarly, attention based architectures~\citep{vaswani_attention} encode prior knowledge about semantic understanding in natural languages. A whole new field --- \textit{geometric deep learning} --- looks at encoding symmetry contraints of the underlying domain into the network architecture~\citep{cohen_equivariant_cnns,gdl_book,gdl_sample_complexity}. Incorporating these priors have shown significant improvements in accuracy as well as sample complexity (number of data points needed for achieving a given accuracy).

Given the importance of incorporating prior knowledge into learning, and the strong prior provided by the manifold hypothesis, multiple methods have been proposed to incorporate this geometric knowledge into learners, either explicitly or implicitly. Traditional machine learning saw a collection of very successful methods, termed as \textit{manifold learning}~\citep{tenenbaum2000isomap_nonlinear_dim_reduction,roweis2000nonlinear_LLE,zhang2004tangent_space_alignment}, which aimed to directly learn the manifold. For incorporating manifold prior into modern DNNs, the most popular and commonly used method is domain aware augmentation, where new samples are generated via manifold preserving transformations known from domain knowledge (e.g. translation, scaling, rotation, etc. for image classification). Other methods such as Manifold Tangent Classifier~\citep{rifai2011manifold_tangent_classifier} which encourage network invariance along learnt tangent spaces have also been proposed. However, other than augmentations, such methods have seen limited adoption.

In this paper, we propose a new method, \distlearner{}, to incorporate the manifold prior for DNN-based classifiers. While standard classifiers are trained to learn the class label for each input point, \distlearner{} is trained to predict the distance of an input point from the underlying manifold of each class. This is enabled by generating points (similar to augmentation) at a \textit{known} distance to the underlying manifold. We do this by approximating the local manifold near each training point and adding controlled perturbations to the point in both the tangent space and normal space (orthogonal compliment of the tangent space). The \distlearner{} is then trained via an MSE loss to minimize the error between predicted and actual distance of these generated points, as well as the original training points (which have a distance of zero to the manifold). During inference, \distlearner{}, chooses the class corresponding to the closest predicted class manifold.

\distlearner{} has many advantages. First, it can learn much more meaningful decision regions (Figure~\ref{fig:decision_boundary}), as classification is now based on distances to the class manifolds. By providing a distance along with the class label, our method is able to take advantage of the richer geometric information per sample. Second, \distlearner{} can identify points that are out of distribution (belonging to neither class), as their predicted distance to all class manifolds will be very high. Standard classifiers on the other hand, are forced to give an output for \textit{any} input. For example, a standard classifier trained to classify images of cats and dogs, will classify images of unrelated objects, say a car or even random noise, as a cat or a dog. DNNs often make such predictions with a high confidence~\citep{hein2019relu} making identification of these cases hard. Third, learning meaningful boundaries can help make \distlearner{} more robust to adversarial attacks. Vulnerability to adversarial attacks~\citep{goodfellow2014explaining,huang2017adversarial,akhtar2018threat_adversarial,chakraborty2018adversarial,Madry2018TowardsDL} has been attributed by many researchers to the high dimensionality of the representation space as well as that of the underlying data manifold~\citep{gilmer2018adversarial,Stutz2019CVPR}. Providing more structured geometric information about the manifold, can help alleviate these issues.

We perform experiments with multiple synthetic datasets to evaluate \distlearner{}. We first visualize the distance boundaries learnt by \distlearner{} and standard classifier on low-dimensional datasets, for a qualitative evaluation. Our experiments show that \distlearner{} is able to learn much more meaningful boundaries, corresponding to the actual distance to underlying class manifolds. For experiments with high dimensional data, we build on the concentric spheres dataset proposed in~\cite{gilmer2018adversarial}. We evaluate on the task of adversarial robustness, and show that \distlearner{} not only outperforms the standard classifier significantly, but also performs at par with classifiers trained via state-of-the-art adversarial training~\citep{Madry2018TowardsDL}.

The contributions of our work can be summarized as follows:
\begin{enumerate}
\vspace{-10pt}
\itemsep0em 
    \item A new learning paradigm, \distlearner{}, which incorporates geometric information about the underlying data manifold by learning distances from the class manifolds.
    \item An efficient method to generate points at a known distance to the class manifolds, enabled by learning the local manifold near each training point.
    \item Evaluation of the proposed approach on multiple synthetic datasets. In particular, we demonstrate that \distlearner{} has adversarial robustness comparable to adversarial training~\cite{Madry2018TowardsDL}. We plan to open-source our code upon acceptance.
\end{enumerate}

\section{Distance Learner}
\label{sec:distlearn}

The manifold hypothesis states that real world high-dimensional data lies on or near low-dimensional manifolds.
Consider a classification dataset, $\mathcal{D}$, consisting of $C \in \mathbb{Z}^+$ classes:
\begin{equation*}
    \mathcal{D} = \{(\mathbf{x}_i, y_i) | \mathbf{x}_i \in \mathbb{R}^n, y_i \in \mathbb{Z}^+, 1 \leq y_i \leq C\},
\end{equation*}
where $\mathbf{x}_i \in \mathbb{R}^n$ is an input point and $y_i$ is the corresponding class label. Let $\mathcal{D}_c$ be the set of all points in class $c$, i.e. $\mathcal{D}_c = \{\mathbf{x}_i | (\mathbf{x}_i, y_i) \in \mathcal{D}, y_i = c\}$. The manifold hypothesis then states that the set $\mathcal{D}_c$ lies on a topological manifold, say $\mathcal{M}_c$, of dimension $m$ much lower than $n$. Our aim is to incorporate geometric information about these manifolds, $\mathcal{M}_c$, into learning.

We propose to incorporate this geometric information via learning a model which can predict the distance of a given point to the manifold\footnote{Distance of a point $\mathbf{p}$ to a manifold $\mathcal{M}$ is defined as the minimum distance of $\mathbf{p}$ to the any point on $\mathcal{M}$.}. That is, we want the model to learn a function $f(\mathbf{x}) = \mathbf{d}$, where $\mathbf{x}\in\mathbb{R}^n$ and the output $\mathbf{d}\in\mathbb{R}^C$ contains the predicted distance to each class manifold. We call such a learner, which captures the distance field of a manifold, the \distlearner{}. We note that the distance field of a manifold captures the entire manifold information, since one can fully recover the actual manifold as the zero level-set of the distance field. Thus distance learning can be a powerful way of incorporating manifold information.

For training \distlearner{}, we propose a simple augmentation based strategy. For each point in the dataset $\mathcal{D}$, we generate augmented points which are at a \textit{known} distance to the underlying manifold. We discuss our augmentation method in section~\ref{sec:genaug}. Generating these augmented points, however, requires knowledge of the local manifold around each data point, which is not always available. We discuss a strategy to infer the local manifold for a given dataset $\mathcal{D}$ in section~\ref{sec:inferring_manifold}. Finally, \distlearner{} is trained via an MSE loss to minimize the error between predicted and actual distances of the augmented points, as well as the original training points, which are at a distance of zero to the manifold. We discuss this further in section~\ref{sec:dist_learner_training}.

\subsection{Generating Augmentations}
\label{sec:genaug}

\begin{figure}
    \centering
      \begin{subfigure}[b]{0.69\linewidth}
        \includegraphics[width=\linewidth]{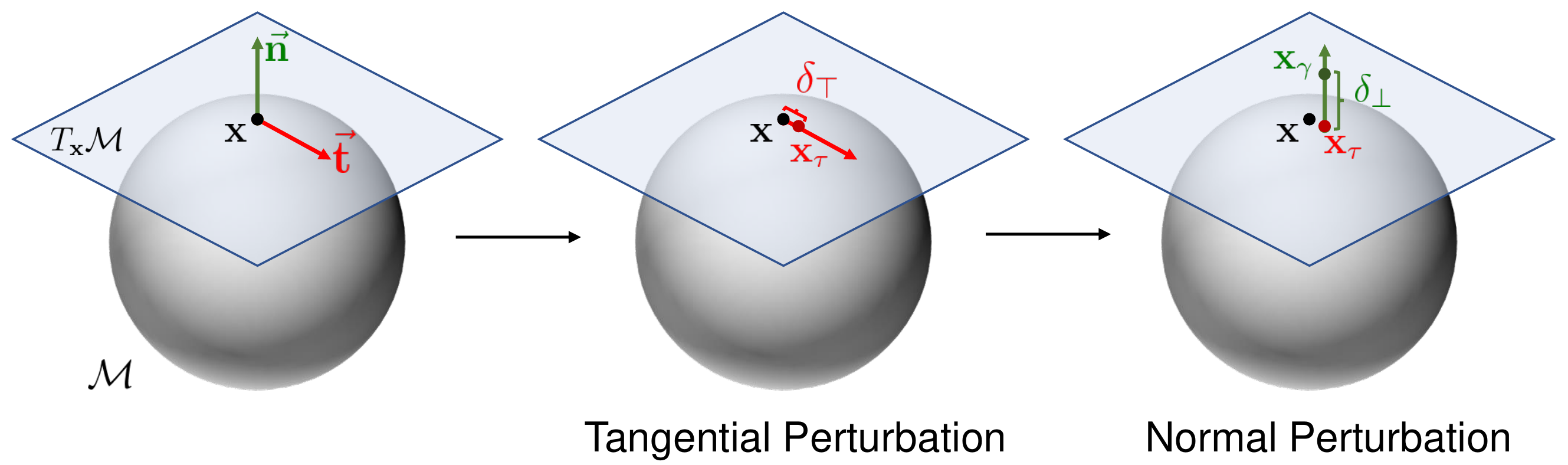}
        \caption{}
        \label{fig:offmfldgen}
      \end{subfigure}
      \begin{subfigure}[b]{0.3\linewidth}
        \includegraphics[width=\linewidth,trim=0mm 0mm 0mm 5.5mm]{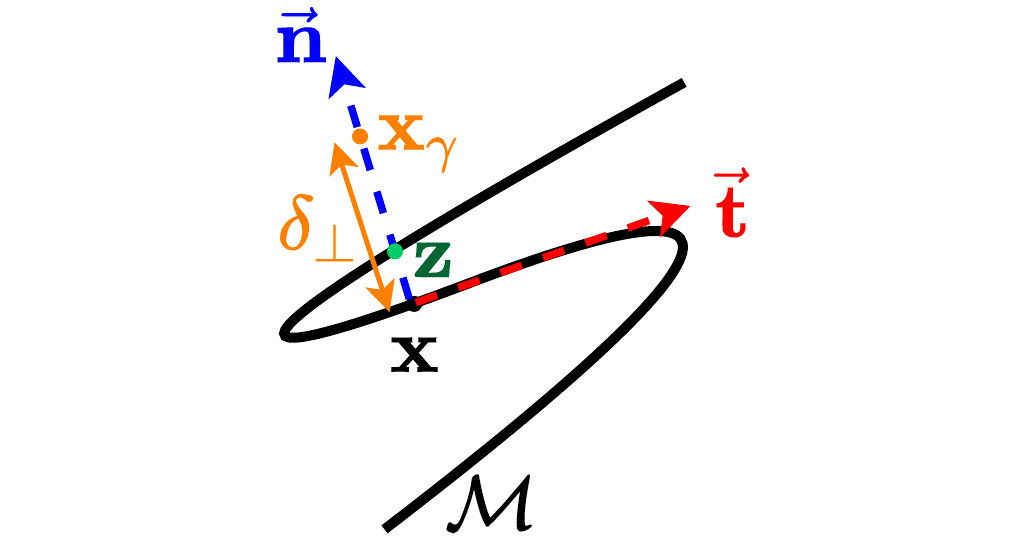}
        \caption{}
        \label{fig:offmfldgenfailure}
      \end{subfigure}
      \caption{\textbf{(a) }Computing off-manifold augmentations: A point $\mathbf{x} \in \mathcal{M}$, is first perturbed in the tangent space $T_\mathbf{x}\mathcal{M}$ by a magnitude of $\delta_\top$ to obtain $\mathbf{x}_\tau$. Next, a perturbation of magnitude $\delta_\perp$ is added in the normal space, $N_\mathbf{x}\mathcal{M}$, to obtain an off-manifold point $\mathbf{x}_\gamma$. The distance of $\mathbf{x}_\gamma$ from $\mathcal{M}$ is then $\sim\delta_\perp$ \textbf{(b)} Example of failure case due to high manifold curvature. The closest manifold point to $\mathbf{x}_\gamma$ is now $\mathbf{z}$, which is at a distance of less than $\delta_\perp$.}
      \label{fig:1}
    
\end{figure}

We need to generate augmented points which are at a known distance to the underlying class manifold. Na\"ively generating points via random sampling will be very expensive as computing distance to class manifolds will require brute-force comparison to all points in $\mathcal{D}$. Moreover, such a method will be very sample inefficient --- we want to sample points close to class manifolds to learn fine-resolution distance fields near the manifolds, but randomly sampled points will be uniformly distributed in the entire domain of interest. Here, we propose an efficient sampling method which generates sample points close to the manifold, and at a known distance. For this, we take a point $\mathbf{x}$ from the training set $\mathcal{D}$ (points in $\mathcal{D}$ are assumed to be on the class manifold), and add a small perturbation to it. To perturb the point to a given distance from the manifold, we will need information about the manifold locally near $\mathbf{x}$. For ease of exposition, we will assume in this section that this information is available. This assumption will be relaxed in section~\ref{sec:inferring_manifold}.

Let the underlying data manifold corresponding to $\mathbf{x}$ be $\mathcal{M}$ of dimensionality $m$. The manifold can be thought of as embedded in $\mathbb{R}^n$. The tangent space $T_{\mathbf{x}}\mathcal{M}$ at the point $\mathbf{x} \in \mathcal{M}$ is then informally defined as the $m$-hyperplane passing through $\mathbf{x}$ and tangent to $\mathcal{M}$. We assume that $T_{\mathbf{x}}\mathcal{M}$ is known, along with an orthonormal spanning basis $BT_\mathbf{x}$. Let $N_{\mathbf{x}}\mathcal{M}$ be the $(n-m)$-dimensional \textit{normal}-space at $\mathbf{x}$, defined as the orthogonal compliment of the tangent space w.r.t the embedded space. Let $BN_\mathbf{x}$ be a corresponding orthonormal spanning basis.

Once the tangent and normal spaces of the manifold near $\mathbf{x}$ are known, we generate an augmented point as follows. We first sample two random vectors $\vec{\mathbf{t}}$ and $\vec{\mathbf{n}}$ from the tangent and normal spaces, respectively. For sampling $\vec{\mathbf{t}}$, we simply sample $m$ scalars uniformly at random from $[0,1]$ and use them as coefficients of the basis $BT_\mathbf{x}$. Sampling $\vec{\mathbf{n}}$ proceeds similarly. The tangent and normal vectors are then combined to generate an augmented point via:
\begin{equation}
\label{eq:off_mfd_generation}
    \mathbf{x}_{\gamma} = \mathbf{x} + \delta_\top\frac{\vec{\mathbf{t}}}{\lVert \vec{\mathbf{t}}\rVert} + \delta_\perp\frac{\vec{\mathbf{n}}}{\lVert \vec{\mathbf{n}}\rVert}.
\end{equation}
That is, we perturb the on manifold point $\mathbf{x}$ in the tangent space by an amount $\delta_\top$ and in the normal space by an amount $\delta_\top$. $\delta_\top$ and $\delta_\perp$ are sampled uniformly at random from $[0, max\_tangent]$ and $[0, max\_norm]$, where $max\_tangent$ and $max\_norm$ are hyperparameters.

\begin{figure}
  \centering
  \begin{subfigure}[b]{0.31\linewidth}
    \includegraphics[width=0.8\linewidth]{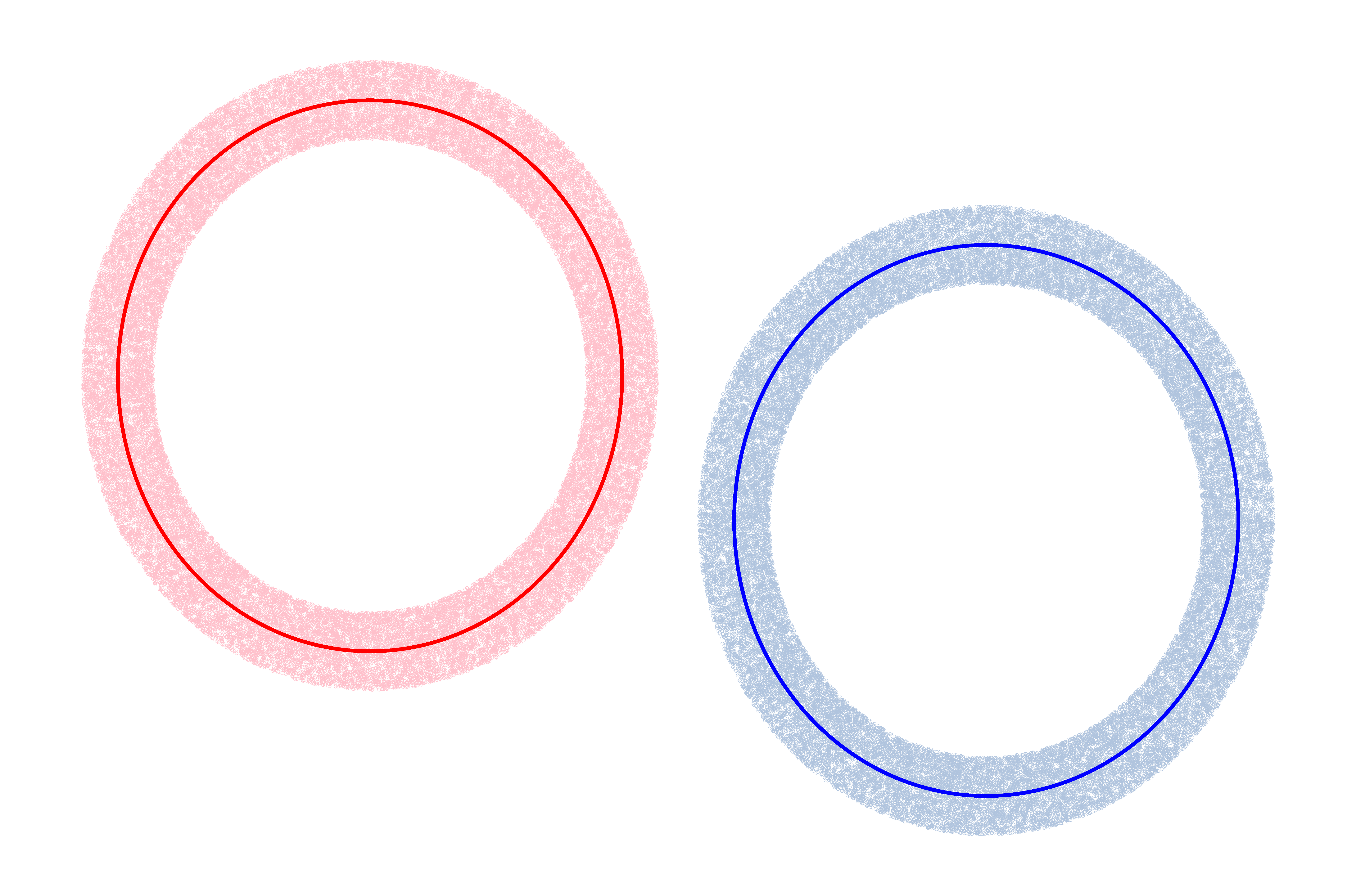}
    \caption{Separated Spheres}
  \end{subfigure} \hfill
  \begin{subfigure}[b]{0.24\linewidth}
    \includegraphics[width=0.8\linewidth]{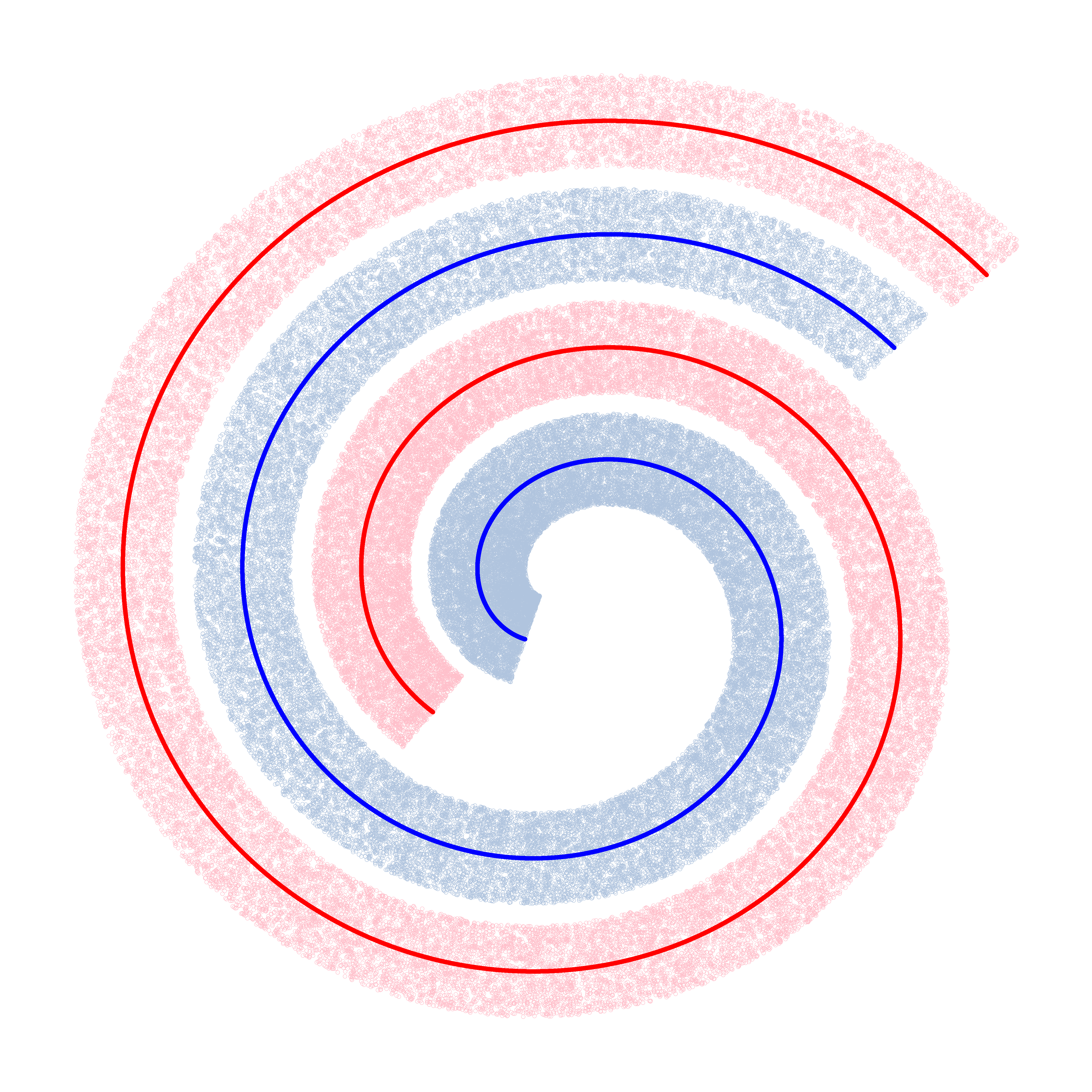}
    \caption{Intertwined Swiss Rolls}
  \end{subfigure} \hfill
  \begin{subfigure}[b]{0.24\linewidth}
    \includegraphics[width=0.8\linewidth]{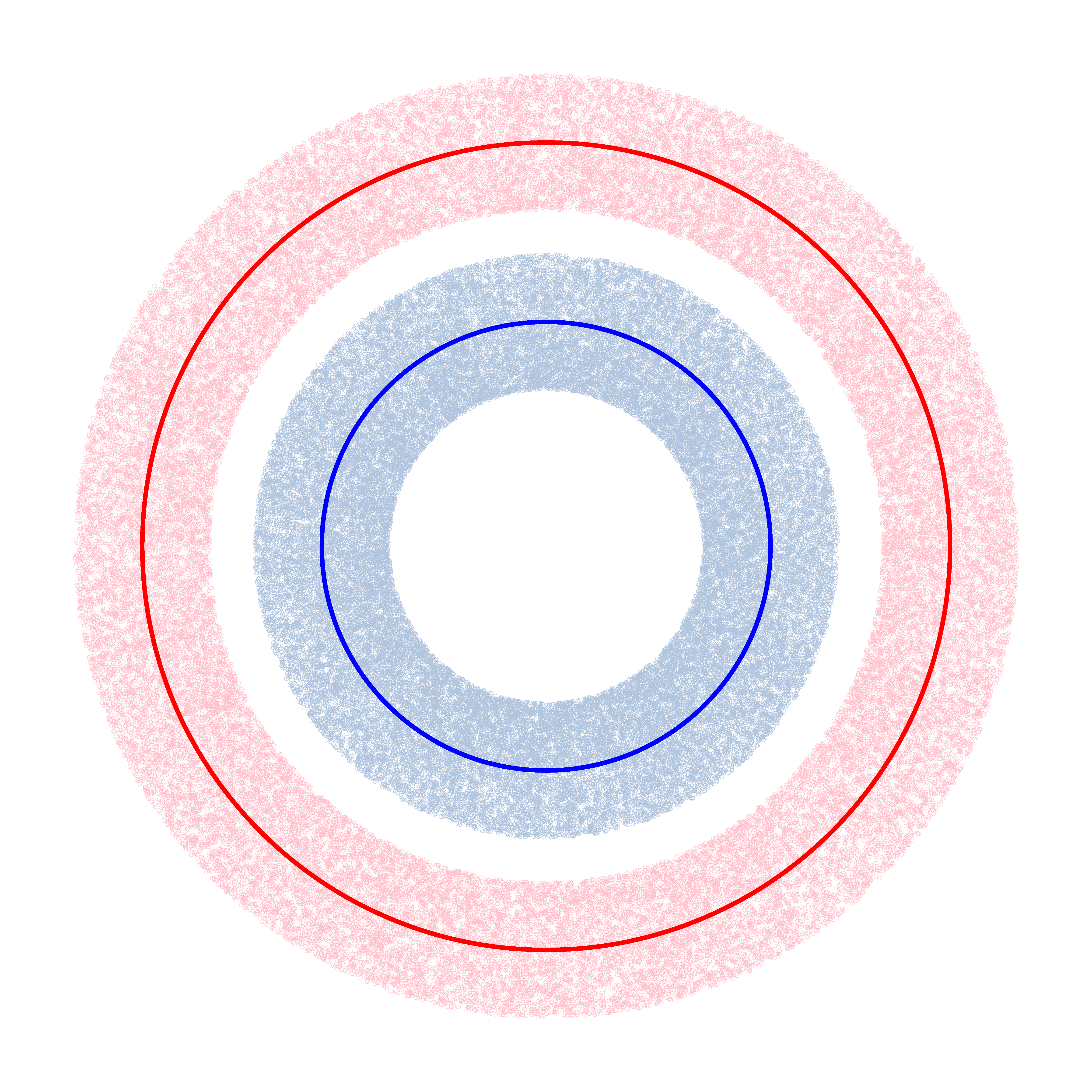}
    \caption{Concentric Spheres}
  \end{subfigure}
  \caption{Synthetic datasets: Representative plots with manifold dimension $m=1$, and embedded dimension $n=2$. Dark blue and dark red points are points on the manifold, while light blue and pink points are the augmented points generated to train \distlearner{}.}
  \label{fig:datasets_and_agumentations}
\end{figure}

If we assume $\delta_\top$, $\delta_\perp$ to be reasonably small, then the distance of $\mathbf{x}_{\gamma}$ from manifold $\mathcal{M}$ can be approximated to be $\delta_\perp$ (see Figure~\ref{fig:offmfldgen}). Since the tangential perturbation $\delta_\top$ is small, the tangentially perturbed point $\mathbf{x}_{\tau} = \mathbf{x} + \delta_\top\frac{\vec{\mathbf{t}}}{\lVert \vec{\mathbf{t}}\rVert}$ can be assumed to stay on the manifold. Also, we can assume that the tangent space and thus the normal space at $\mathbf{x}_{\tau}$ is the same as that at $\mathbf{x}$. Then, since the augmented point $\mathbf{x}_{\gamma}$ is generated by adding a small perturbation perpendicular to the tangent space at $\mathbf{x}_{\tau}$, the closest on-manifold point to $\mathbf{x}_{\gamma}$ will be $\mathbf{x}_{\tau}$, and thus the distance of the augmented point from the manifold will be $\lVert\mathbf{x}_{\gamma}-\mathbf{x}_{\tau}\rVert = \delta_\perp$. Note, however, that if the manifold has a very high curvature relative to $\delta_\top$ and $\delta_\perp$, this assumption may break (see Figure~\ref{fig:offmfldgenfailure}). Thus, $max\_tangent$ and $max\_norm$ need to chosen reasonably. Figure~\ref{fig:datasets_and_agumentations} gives a few examples of the generated points.

To summarize, given a point $\mathbf{x} \in \mathcal{D}$ and the tangent, normal spaces of the class manifold $\mathcal{M}$ at $\mathbf{x}$, we can generate an augmented point at a distance of $\delta_\perp$ from the manifold $\mathcal{M}$ using equation~\ref{eq:off_mfd_generation}. Multiple augmented points can be generated for each $\mathbf{x}$ by random sampling of $\vec{\mathbf{t}}, \vec{\mathbf{n}}, \delta_\top$ and $\delta_\perp$.

\subsection{Inferring the Local Manifold}
\label{sec:inferring_manifold}

In section~\ref{sec:genaug}, we assumed availability of the local manifold information near each point $\mathbf{x} \in \mathcal{D}$. In this section we relax this assumption, and discuss how we can infer the local manifold from the dataset $\mathcal{D}$. We use a simple method based on finding $k$ nearest neighbors and principal component analysis~\citep{Jolliffe:1986:pca}, similar to the method in~\cite{LTSA}. Note that \distlearner{} is agnostic to the specific choice of method used to infer the local manifold, and one can use more complex methods, such as the ones discussed in~\citet{lin2008riemannian,rifai2011manifold_tangent_classifier}.

To infer the local manifold near a point $\mathbf{x} \in \mathcal{D}$, we first find the $k$ nearest neighbors (kNNs) of $\mathbf{x}$ in the dataset $\mathcal{D}$. We denote this set of kNNs as $\textsc{NN}_\mathcal{M}(\mathbf{x}, k)$, where $\mathcal{M}$ is the underlying class manifold of $\mathbf{x}$. If the neighbors are close enough, it is reasonable to assume that they will belong to $\mathcal{M}$. Note that for nearest neighbor search, we use the Euclidean distance in the embedded space $\mathbb{R}^n$, and not the geodesic distance on $\mathcal{M}$. As a result, even though the neighbors are close in $\mathbb{R}^n$, they may not be close in $\mathcal{M}$. However, if $\mathcal{M}$ satisfies certain properties (e.g. bounded curvature), we can assume closeness w.r.t geodesic distance as well. See~\citet{lin2008riemannian} for a discussion on these properties and failure cases.

Since locally a manifold is isomorphic to $\mathbb{R}^m$ ($m$ is the manifold dimensionality), we can assume the close neighbors, $\textsc{NN}_\mathcal{M}(\mathbf{x}, s)$, to lie on $\mathbb{R}^m$. Thus, $\textsc{NN}_\mathcal{M}(\mathbf{x}, s)$ can be used to construct the tangent space $T_\mathbf{x}\mathcal{M}$ as follows. Let
$
    L_\mathcal{M}(\mathbf{x}) = \{\mathbf{x}_j - \mathbf{x} | \mathbf{x}_j \in \textsc{NN}_\mathcal{M}(\mathbf{x}, k)\}
$.
The tangent space, $T_\mathbf{x}\mathcal{M}$, is then the span of vectors in $L_\mathcal{M}$ (as long as the number of independent vectors in $L_\mathcal{M}$ is $\ge m$).

Because of the discrete nature of $\mathcal{D}$, the nearest neighbors will be a finite distance away from $\mathbf{x}$, and will not lie exactly on the local manifold. Thus, the span of $L_\mathcal{M}$ will include components outside the tangent space, $T_\mathbf{x}\mathcal{M}$. The severity of this problem will depend on the sampling density of $\mathcal{D}$. To remove these noisy components, we do a Principal Component Analysis (PCA) on $L_\mathcal{M}$, and extract the top-$m$ most important principal components (based on their explained variance). These components then form the orthonormal basis, $BT_\mathbf{x}$, of the tangent space $T_\mathbf{x}\mathcal{M}$.

Finally, we need to compute an orthonormal basis $BN_\mathbf{x}$ of the $(n-m)$-dimensional \textit{normal}-space, $N_{\mathbf{x}}\mathcal{M}$, at $\mathbf{x}$. Since all vectors in $N_{\mathbf{x}}\mathcal{M}$ are orthogonal to the vectors in $T_{\mathbf{x}}\mathcal{M}$ (i.e., $\vec{\mathbf{n}} \cdot \vec{\mathbf{t}} = 0$, $\forall \vec{\mathbf{n}} \in N_\mathbf{x}\mathcal{M}, \vec{\mathbf{t}} \in T_\mathbf{x}\mathcal{M}$), the normal-space $N_\mathbf{x}\mathcal{M}$ is the null-space of $T_\mathbf{x}\mathcal{M}$. The basis, $BN_\mathbf{x}$, of this null-space can thus be computed by SVD of a matrix whose rows are elements of $BT_\mathbf{x}$.

Note that here we have assumed knowledge of the intrinsic dimensionality of the manifold $\mathcal{M}$, to choose the number of principal components, as well as to choose the minimum number of neighbors $k$. There are multiple methods to compute the intrinsic dimensionality of a manifold from its nearest neighbors (e.g.,~\citep{lin2008riemannian}) which can be used for the purpose.

\subsection{Learning Distance}
\label{sec:dist_learner_training}

The augmented points generated above, can now used to train our \distlearner{}. Note, however, that our generated point $\mathbf{x}_{\gamma}$ has its distance known to only \textit{one} class manifold --- the manifold on which the point $\mathbf{x}$ lies. However, we want our \distlearner{} to predict distances to \textit{all} class manifolds. To enable this, we simply set the distance of the generated point to all other-class manifolds at a fixed high value, $high\_distance$. Similarly points in the dataset $\mathcal{D}$ have their distance set to zero for the class manifold they belong to, and to $high\_distance$ for all other classes. Since the different class manifolds can be assumed to be well separated, points on or near one manifold will be quite far from all other manifolds. Hence, setting a high value for distances to other-class manifolds captures the coarse distance information. 

Also note that since the normal perturbation $\delta_\perp \in [0, max\_norm]$, our augmented points are only generated in $max\_norm$ thick bands around each class manifold (see Figure~\ref{fig:datasets_and_agumentations}). However, since these bands contain fine distance information, it is enough to encode the manifold structure (the manifold can still be recovered as the zero level-set). Moreover, the other-class manifolds provide samples at a high distance, which encourage the model to predict high values for points far outside these bands. In our experiments, we observe that even outside these bands, \distlearner{} is able to learn a smoothly increasing function (Figure~\ref{fig:dist_heatmap}). We note that it is a common practice in level-sets based surface representation methods (e.g. liquid surface representation for simulating fluids on a grid), where only small bands around the surface store the actual signed distance function~\citep{osher1988fronts,osher2001level,enright2002hybrid}.

Our method is agnostic to the specific architecture of the model. The model should have $C$ outputs, corresponding to the predicted distance to each of the $C$ classes. We train the model by minimizing the Mean Squared Error between the predicted and ground-truth distances.

\subsection{Classification using Distance Learner}
\label{sec:classification_dist_learner}

For classification using \distlearner{}, we use two variants. In the first variant, we classify input points as belonging to one of the $C$ classes. This is done by outputting the class with the closest predicted class manifold. In the second variant, we also allow classifying points as being out-of-domain (i.e., belonging to neither class), if the distance to the closest manifold is higher than a tolerance threshold, say $tol$. The tolerance corresponds to the distance from the class manifold which is considered close enough to still belong to that class. It can be determined from domain knowledge or empirically from a validation set. Since \distlearner{} learns fine-grained distances only in $max\_norm$ bands around the class manifolds, we should choose $max\_norm >  tol$ during training, so that \distlearner{} has high fidelity predictions in the tolerance region. Note that standard classifiers have to classify all points (even completely unrelated points far from any class manifold) to one of the $C$ classes the classifier was trained on. \distlearner{} does not have this limitation.

\section{Experiments}
\label{sec:experiments}

We evaluate \distlearner{} on multiple synthetic datasets and tasks. First, we do a set of qualitative evaluations to visualize the distance function and decision boundaries learnt by the \distlearner{}. To allow visualization, we stick to manifolds of 1 and 2 dimensions in these experiments. We embed the manifolds in much higher dimensions, and visualize 2D slices. This is followed by quantitative experiments to evaluate the adversarial robustness of \distlearner{}.

\noindent\textbf{Model Architecture:} Our \distlearner{} is based on a simple MLP-based architecture, inspired by recent works on learning neural implicit fields for 3D shapes~\citep{deepsdf}. Figure~\ref{fig:model} gives a detailed description of the architecture. The model is trained to minimize an MSE loss. We compare \distlearner{} against a \stdclf{} baseline. Our \stdclf{} uses the same architecture as \distlearner{}, except we apply a softmax layer on the combined output of the two branches. The \stdclf{} is trained via a cross-entropy loss. An Adam optimizer~\citep{adam} is used for both models.

\subsection{Datasets}
\label{sec:datasets}

We evaluate on three sets of synthetic data --- separated spheres, intertwined swiss rolls and concentric spheres (see Figure~\ref{fig:datasets_and_agumentations}). In each, $m$-dimensional manifolds are embedded in $\mathbb{R}^n$ ($m < n$). We experiment with multiple values of $m$ and $n$ for each dataset. Details of dataset generation and our embedding method is discussed in section~\ref{sec:appendix-data-gen}.

\noindent\textbf{Separated Spheres:} This is a simple dataset consisting of two hyperspheres of the same radius $r=1$, where each sphere belongs to a different class (Figure \ref{fig:datasets_and_agumentations}(a)). The sphere centers are separated by a distance of $2.5$ so that there is no overlap.

\noindent\textbf{Intertwined Swiss Rolls:} Swiss roll~\citep{swiss_roll_dataset_original} is a commonly used benchmark dataset for manifold learning techniques. A 2D Swiss roll embedded in $\mathbb{R}^{3}$ can be parameterized as: $\mathbf{x}(\phi, \psi) = (\phi\sin\phi,\, \phi\cos\phi,\, \psi)$. This can be generalized for an $m$ dimensional Swiss roll embedded in $\mathbb{R}^{m+1}$ as, $\mathbf{x}(\phi, \psi_1, \psi_2, \ldots \psi_{m-1}) = (\phi\sin\phi, \, \phi\cos\phi,  \, \psi_1, \, \psi_2, \, \ldots, \, \psi_{m - 1})$. Embedding in dimensions higher than $m+1$ and other details are discussed in section~\ref{sec:appendix-data-gen}. Our complete dataset consists of two intertwined Swiss rolls (Figure \ref{fig:datasets_and_agumentations}(b)), where each Swiss roll corresponds to a different class.

\noindent\textbf{Concentric Spheres:} This dataset consists of two concentric hyperspheres, with each sphere belonging to a different class (Figure \ref{fig:datasets_and_agumentations}(c)). The dataset is based on~\citet{gilmer2018adversarial}, where they used this dataset extensively to analyze the connection between adversarial examples and high dimensional geometry. As in~\citet{gilmer2018adversarial}, the inner and outer spheres' radii are set to $1.0$ and $1.3$, respectively. However, unlike~\citet{gilmer2018adversarial}, where they used $m = 499,\ n = 500$, we used settings with $m$ much lower than $n$. We do this to more closely model real-world datasets where the intrinsic dimensionality of the underlying data is much smaller than the embedded space dimension. In section~\ref{sec:adversarial_robustness}, we focus on the case $m=50$. This is motivated by the findings of~\citet{intrinsic_dimension}, where for all popular image datasets (e.g. ImageNet), their highest estimate of intrinsic dimensionality was less than $50$. Experiments with other values of $m$ are discussed in section~\ref{sec:appenpdix_adversarial_robustness}. We use $n=500$ in all experiments.

\noindent\textbf{A note on accuracy:} In a test set of 20 million samples, ~\citet{gilmer2018adversarial} observed no errors on the concentric spheres dataset. As a result, the true error rate of the model was unknown, and only a statistical upper bound was available. We observed similar behavior for all our datasets with both \distlearner{} and \stdclf{}. However, in spite of this, we find adversarial examples, similar to~\citet{gilmer2018adversarial}. Thus, we focus on adversarial robustness for quantitative evaluation.

\subsection{Distance Prediction Accuracy}
\label{sec:dist_pred_acc}
\begin{figure}
  \centering
  \begin{subfigure}[b]{0.32\linewidth}
    \includegraphics[width=\linewidth]{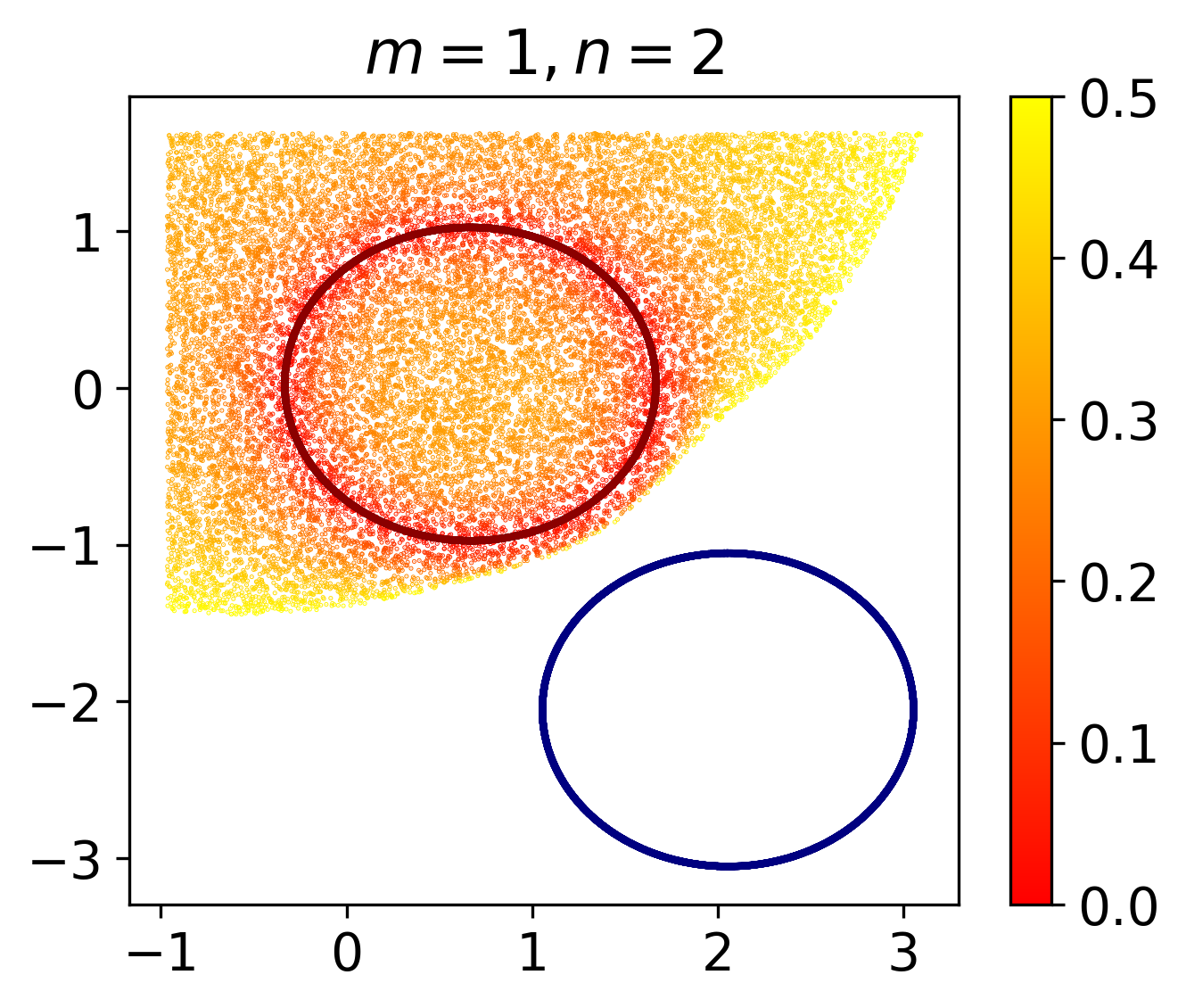}
    \caption{Separated Spheres}
  \end{subfigure}
  \begin{subfigure}[b]{0.32\linewidth}
    \includegraphics[width=\linewidth]{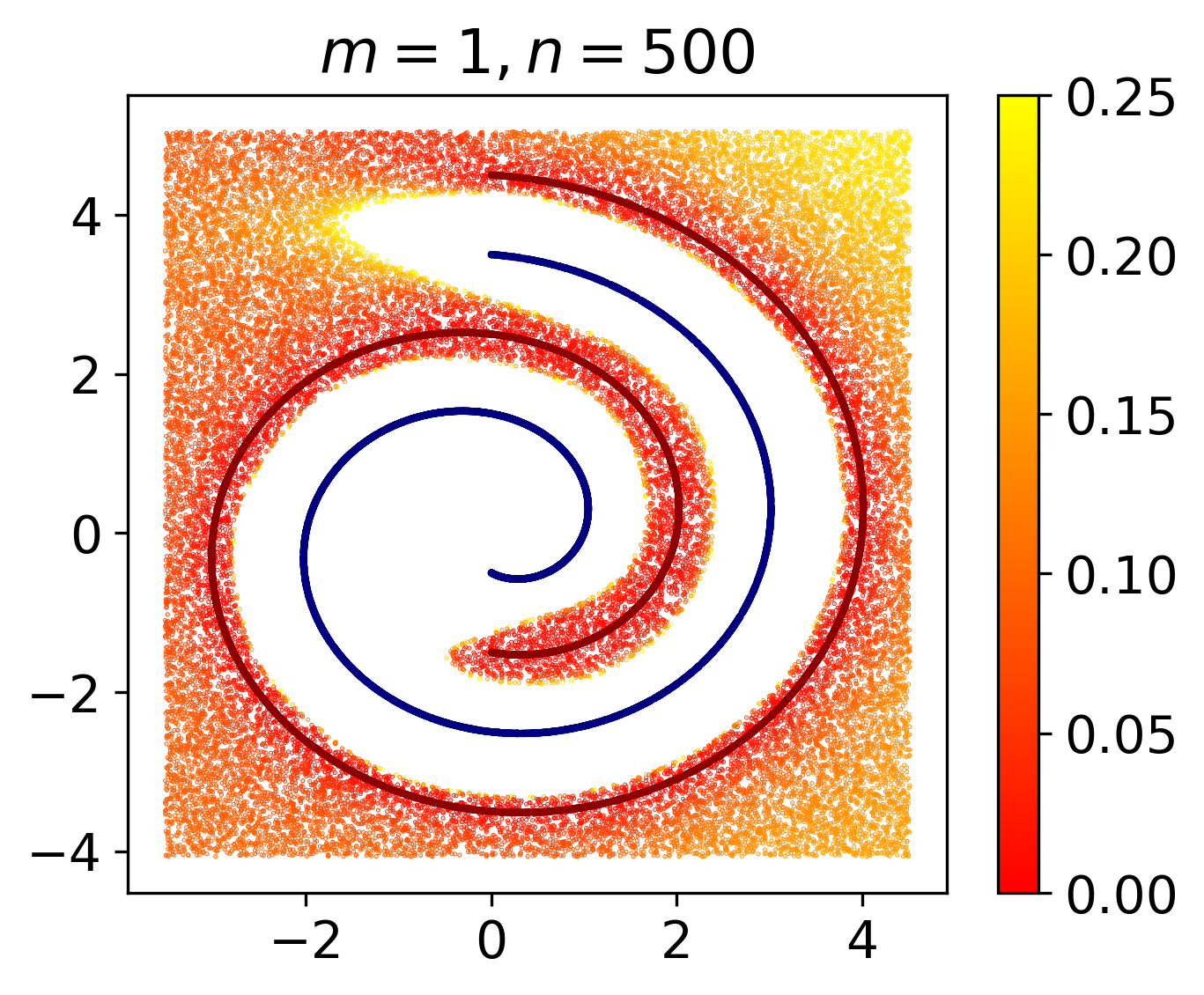}
    \caption{Swiss-Roll}
  \end{subfigure}
  \begin{subfigure}[b]{0.32\linewidth}
    \includegraphics[width=\linewidth]{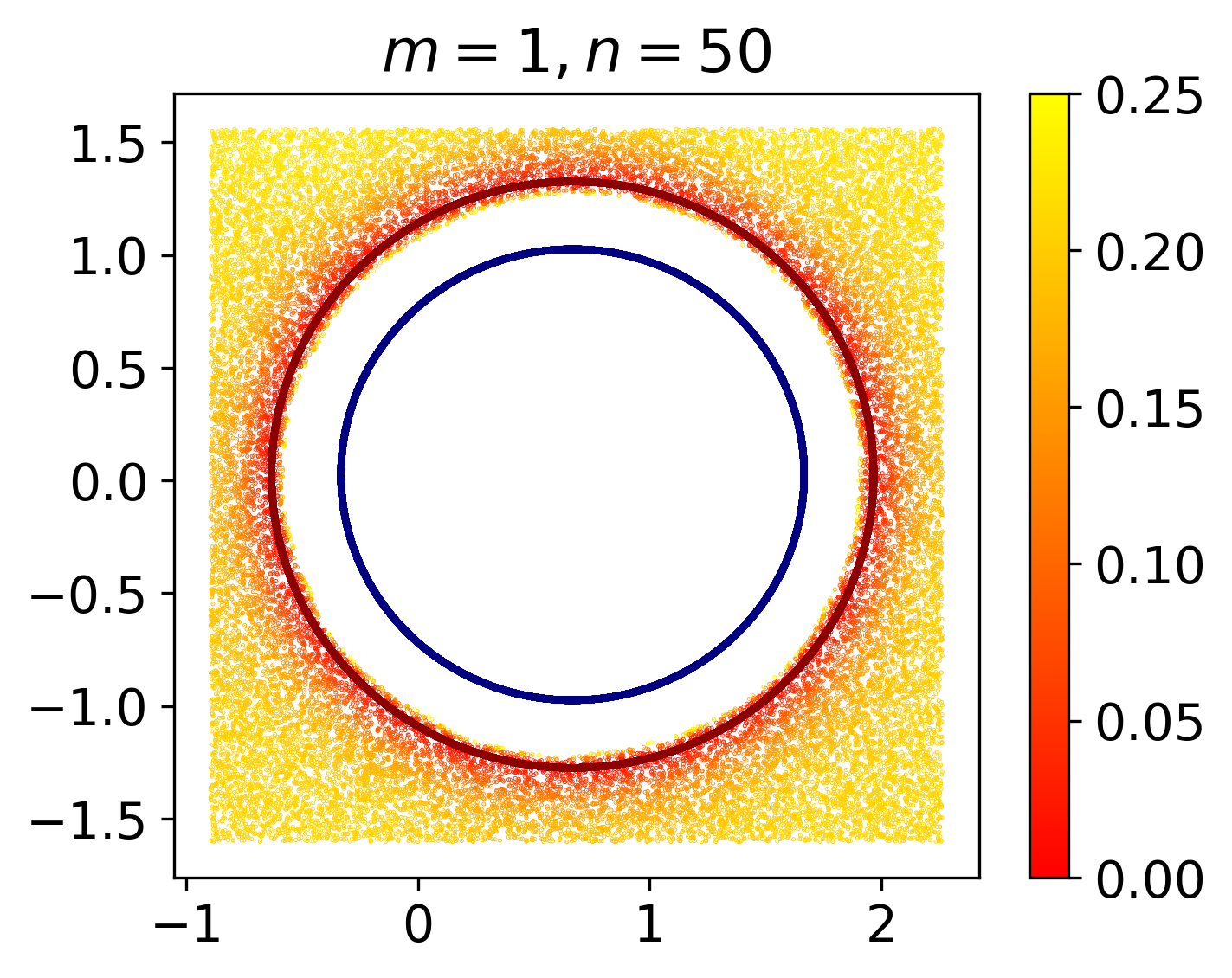}
    \caption{Concentric Spheres}
  \end{subfigure}

  \caption{Learnt distances: Heatmap of predicted distance from the red class manifold is shown in each plot. \distlearner{} is able to accurately predict low distances near the manifold. Interestingly, even though there are no training points far from the manifold, \distlearner{} learns a smoothly increasing function even in the far regions (e.g. see the plot boundaries). The observations hold even when the embedded dimension is very high ($n=500$ in (b)). Note that since the other-class manifold points are trained to have a $high\_distance$ value ($1.0$ in these experiments), the predicted values near the blue class manifolds are close to $1.0$. To enable enough color resolution in the heatmaps, we remove points with predicted distance of $>.5$ in (a) and $>.25$ in (b,c). These points correspond to the blank white region. More detailed plots of the removed region can be found in Figure~\ref{fig:distance_hmap_supple})}
  \label{fig:dist_heatmap}
\end{figure}

\distlearner{} is able to predict distances of points to the class manifolds with high accuracy. Table~\ref{tab:dist_learner_accuracy_3runs} shows that even for high dimensions of manifold ($m$) and embedding space ($n$), the \distlearner{} is able to achieve very low test losses. The mean and standard deviation of the losses are computed over three runs started with difference random seeds. Note that we only use points sampled on the manifold itself and in $max\_norm$ bands around the manifold to compute these losses (full details in section~\ref{sec:appendix_hparams}). Figure~\ref{fig:dist_heatmap} also shows a visualization of the learnt distances as a heatmap. To allow visualization, we only consider relevant 2D slices. Also, here we include points outside the $max\_norm$ bands. Note that even though there are no points outside the $max\_norm$ bands during training, the network is able to learn a smoothly increasing function for these points.

\begin{table}[!ht]
    \small
    \centering
    \caption{Train and test loss of \distlearner{}}
    \begin{tabular}{ccc|ccc}
    \toprule
        Dataset & $m$ & $n$ & Train Loss (av) & Test Loss (av) & \# Test Examples  \\ \midrule
        \multirow{4}{*}{Separated Spheres}  & 1 & 2 & \loss{2.450}{0.157}{-8} & \loss{2.719}{0.185}{-8} & 20,000  \\ 
         & 1 & 50 & \loss{1.268}{0.049}{-6} & \loss{3.151}{0.265}{-7} & 20,000  \\ 
         & 1 & 500 & \loss{2.287}{0.143}{-6} & \loss{3.151}{0.231}{-7} & 20,000  \\ 
         & 2 & 500 & \loss{2.729}{0.093}{-7} & \loss{2.973}{0.058}{-6} & 20,000  \\ \midrule
        \multirow{2}{*}{Intertwined Swiss Rolls} & 1 & 2 & \loss{3.700}{0.289}{-7} & \loss{3.884}{0.317}{-7} & 20,000  \\
         & 1 & 50 & \loss{9.792}{1.657}{-7} & \loss{2.256}{0.550}{-6} & 20,000  \\
         & 1 & 500 & \loss{9.470}{0.984}{-6} & \loss{1.843}{0.248}{-5} & 20,000  \\ \midrule
        \multirow{4}{*}{Concentric Spheres} & 1 & 1 & \loss{4.757}{0.478}{-9} & \loss{5.114}{0.523}{-9} & 200,000  \\ 
         & 1 & 50 & \loss{2.754}{0.053}{-7} & \loss{3.506}{0.208}{-7} & 200,000  \\
         & 2 & 50 & \loss{5.118}{0.230}{-8} & \loss{1.283}{0.215}{-7} & 200,000  \\
         & 25 & 500 & \loss{1.163}{0.024}{-6} & \loss{2.641}{2.198}{-6} & 200,000  \\
         & 50 & 500 & \loss{1.132}{0.047}{-6} & \loss{1.380}{0.547}{-5} & 200,000  \\ \bottomrule
    \end{tabular}
    \label{tab:dist_learner_accuracy_3runs}
\end{table}

\subsection{Out-of-domain Classification \& Decision Boundaries}

\begin{figure}[t]
  \centering
  \begingroup
  \setlength{\tabcolsep}{4pt}
  \begin{tabular}{ccc}
    \includegraphics[width=0.31\linewidth]{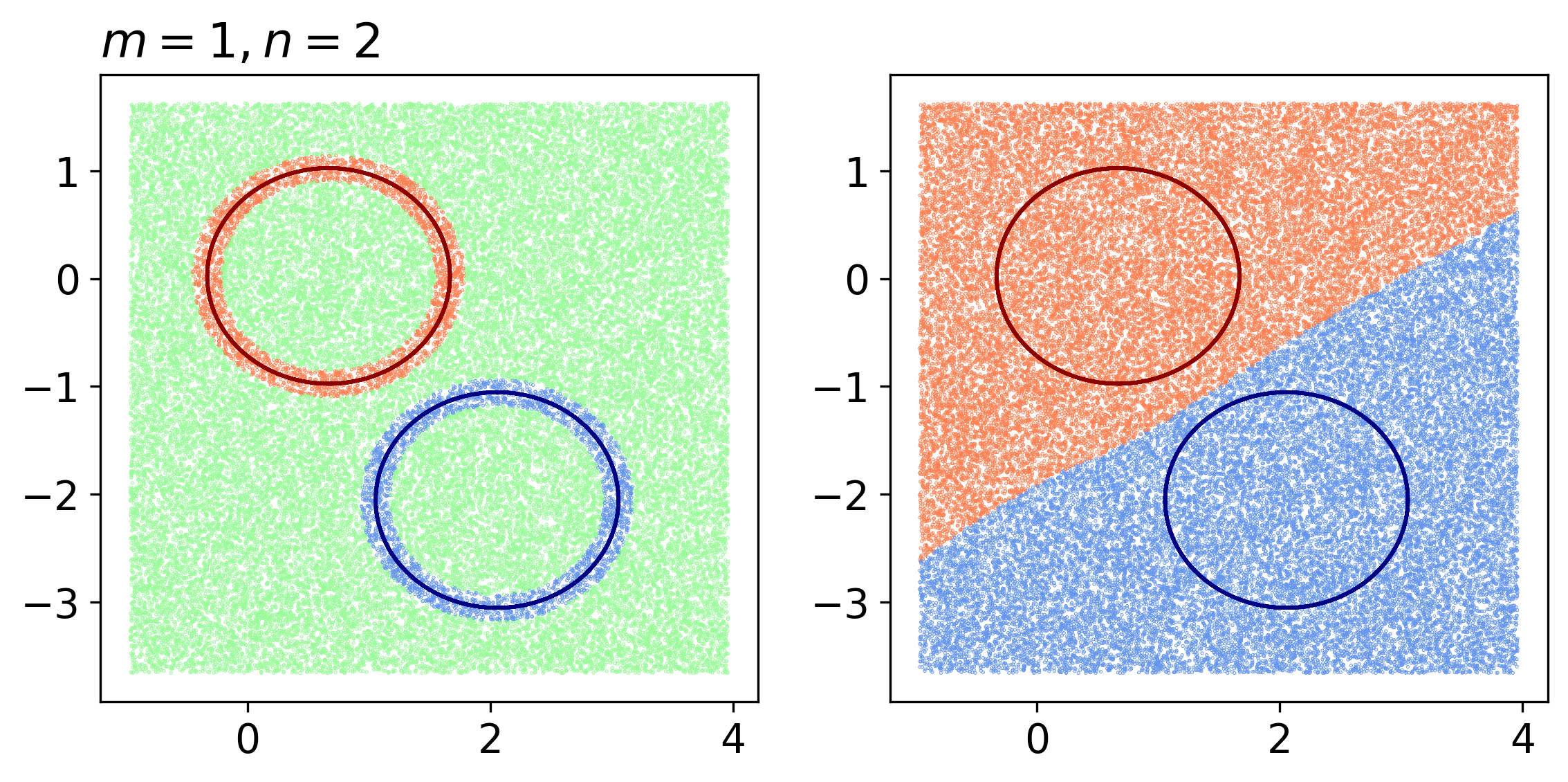} &
    \includegraphics[width=0.31\linewidth]{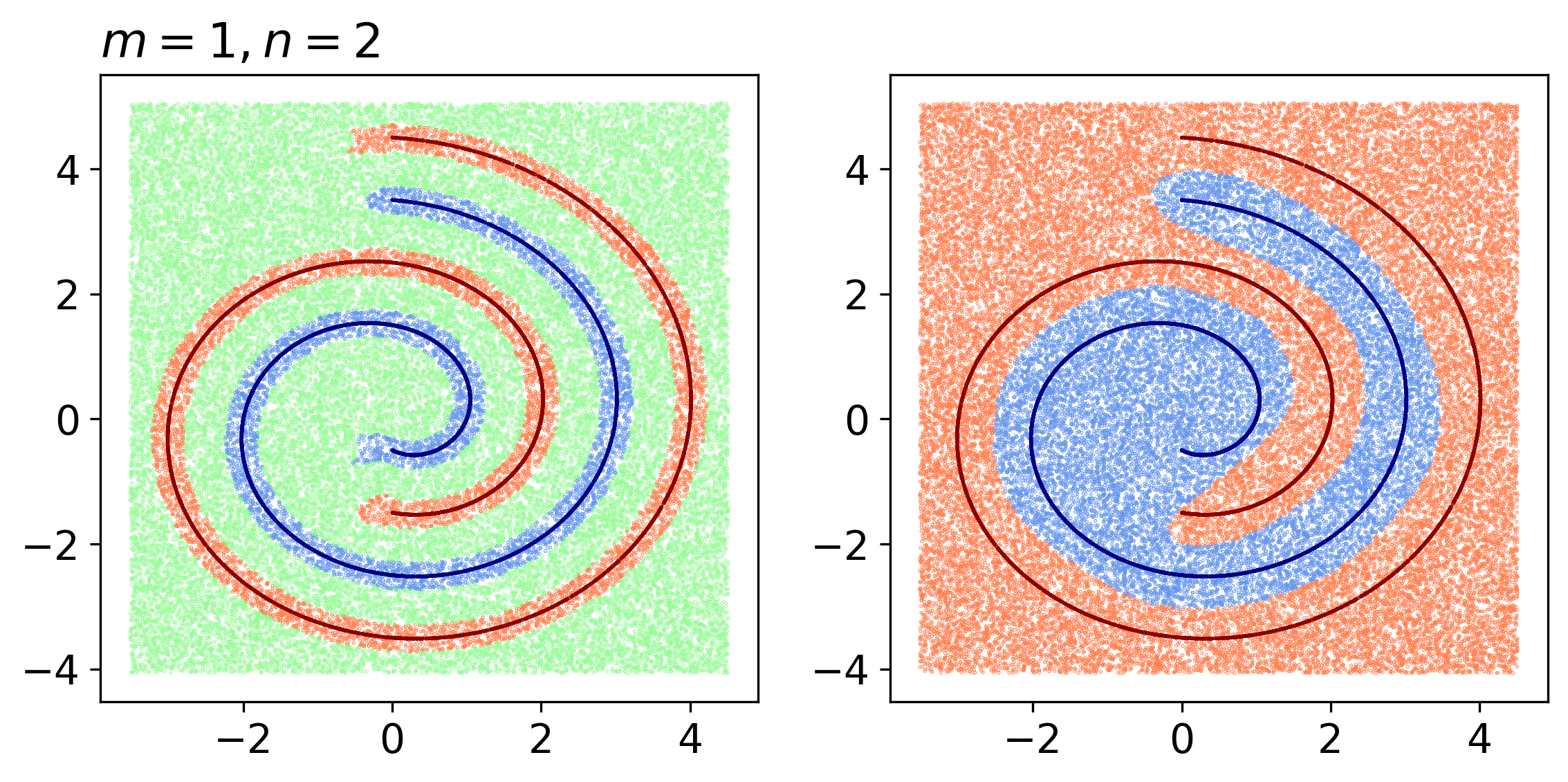} &
    \includegraphics[width=0.31\linewidth]{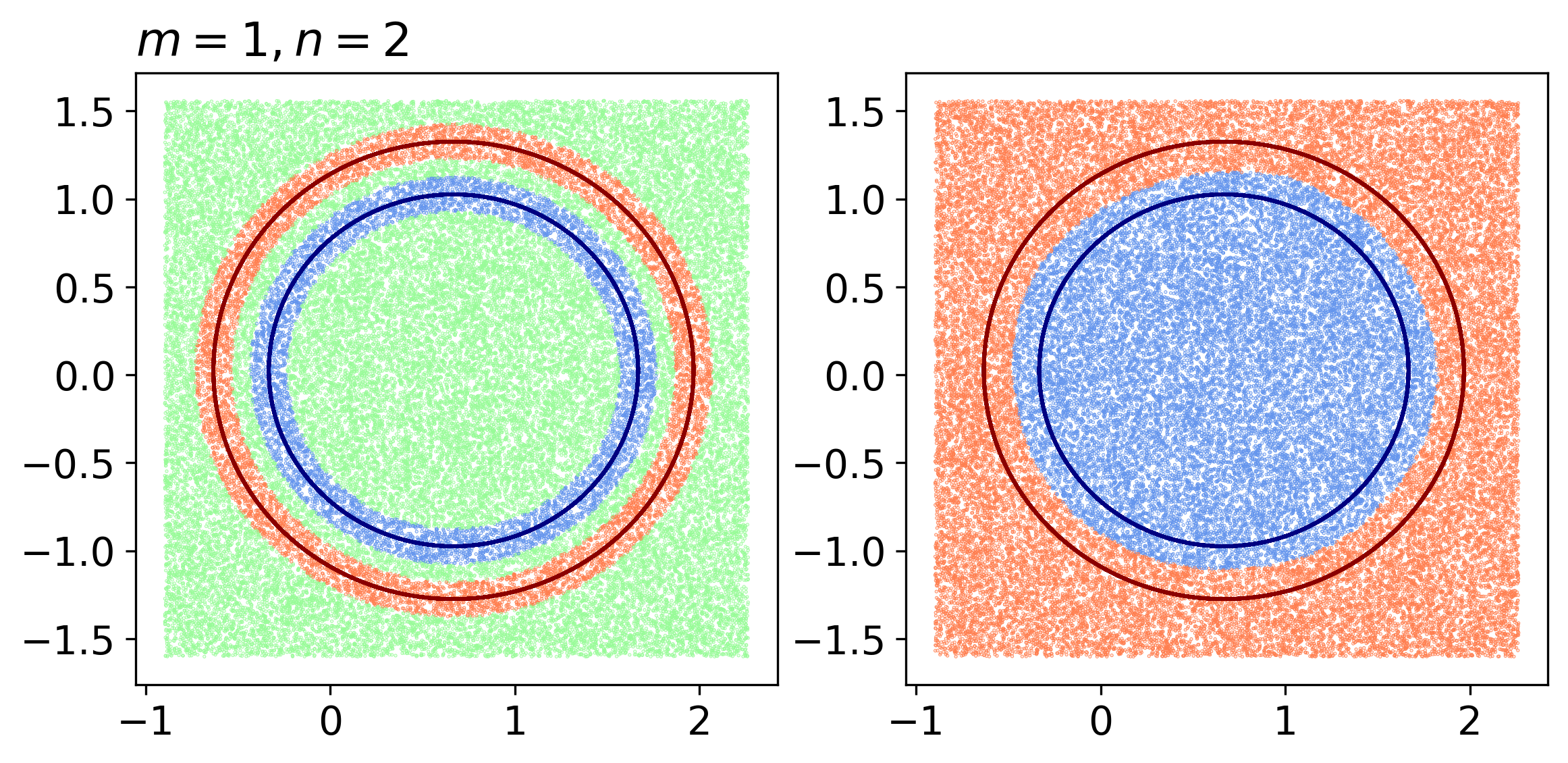}
    \\
    \includegraphics[width=0.31\linewidth]{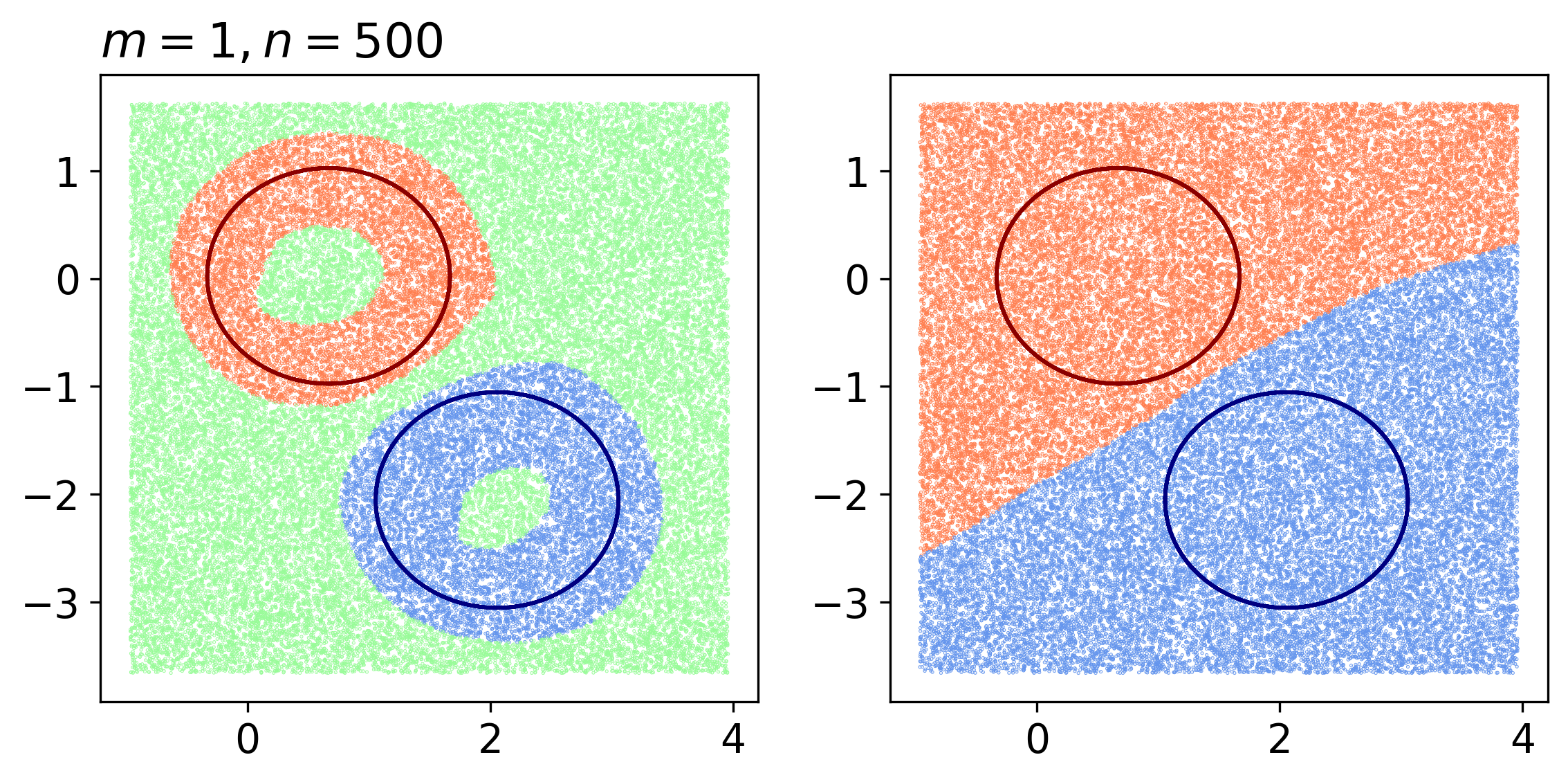} &
    \includegraphics[width=0.31\linewidth]{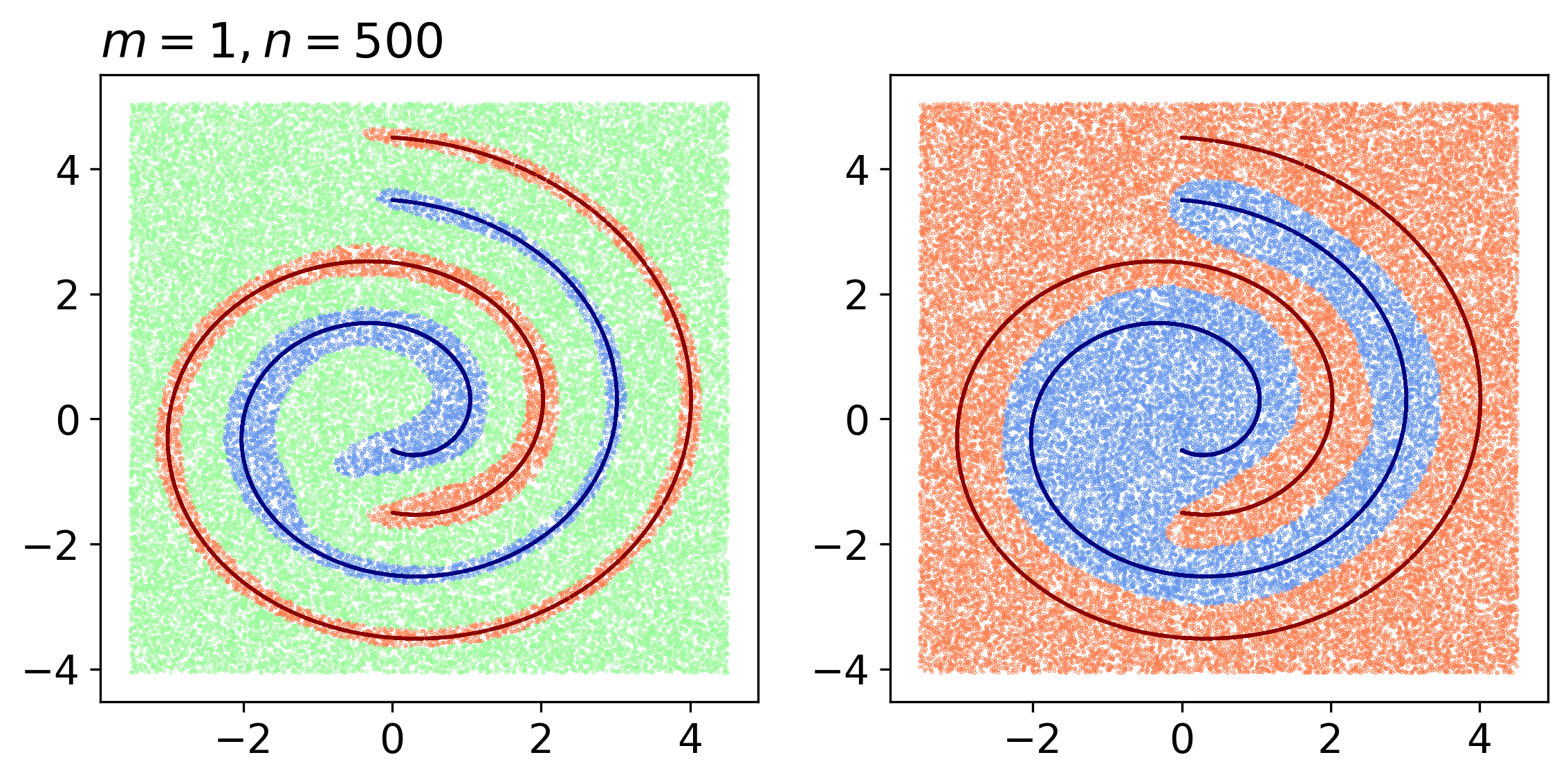} &
    \includegraphics[width=0.31\linewidth]{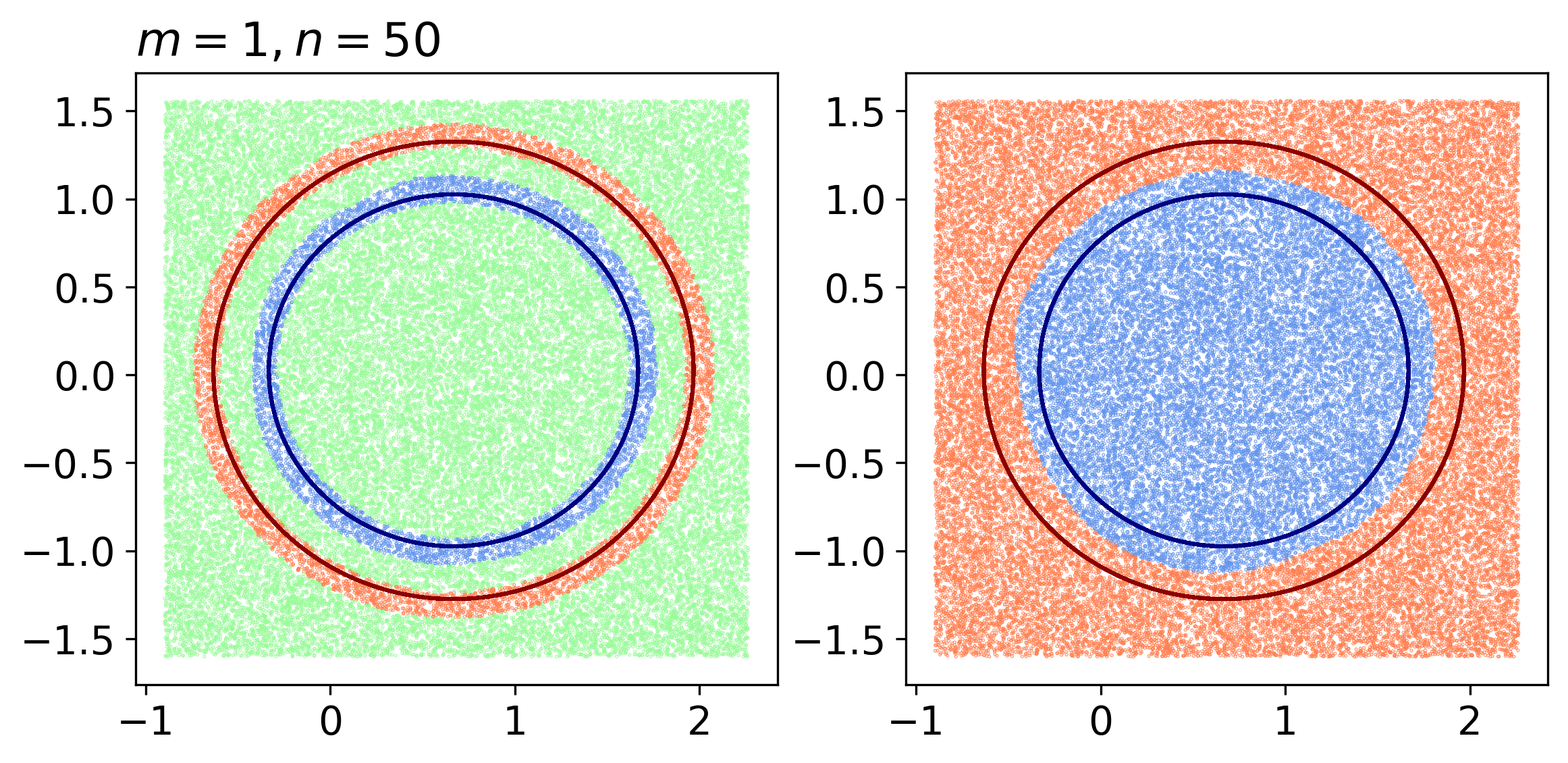}
    \\
    {(a) Separated Spheres} & {(b) Swiss Roll} & { (c) Concentric Spheres}
  \end{tabular}
  \endgroup
  \caption{Decision Regions: Side-by-side comparison of decision regions for the \distlearner{} (left) and \stdclf{} (right). For easy visualization, only manifold dimensions of $m=1$ are used. The two rows correspond to experiments with different values of embedding dimension $n$. Dark solid lines correspond to the actual class-manifolds, while dots correspond to samples in the test set. The blue and red dots are color coded with the color of the predicted class, while green dots in the \distlearner{} plots corresponds to points identified as out-of-domain. While the decision regions learnt by \stdclf{} are very coarse and include even points very far from the manifold, \distlearner{} is able to learn fine regions containing only points close the class manifolds. The observation holds true even for both low and high dimensions.}
  \label{fig:decision_boundary}
\end{figure}

Since \distlearner{} is able to predict distances to the class manifolds with good accuracy, we can use it to identify points that are out-of-domain (i.e., belonging to neither class). As discussed in section~\ref{sec:classification_dist_learner}, points whose predicted distance to the closest manifold is higher than a tolerance threshold, $tol$, are classified as out-of-domain. Figure~\ref{fig:decision_boundary} shows a visualization (over 2D slices) of the classification done by \stdclf{} and \distlearner{}.

\begin{figure}
\vspace{-12pt}
  \centering
  \begin{subfigure}[b]{0.3\linewidth}
    \includegraphics[width=\linewidth]{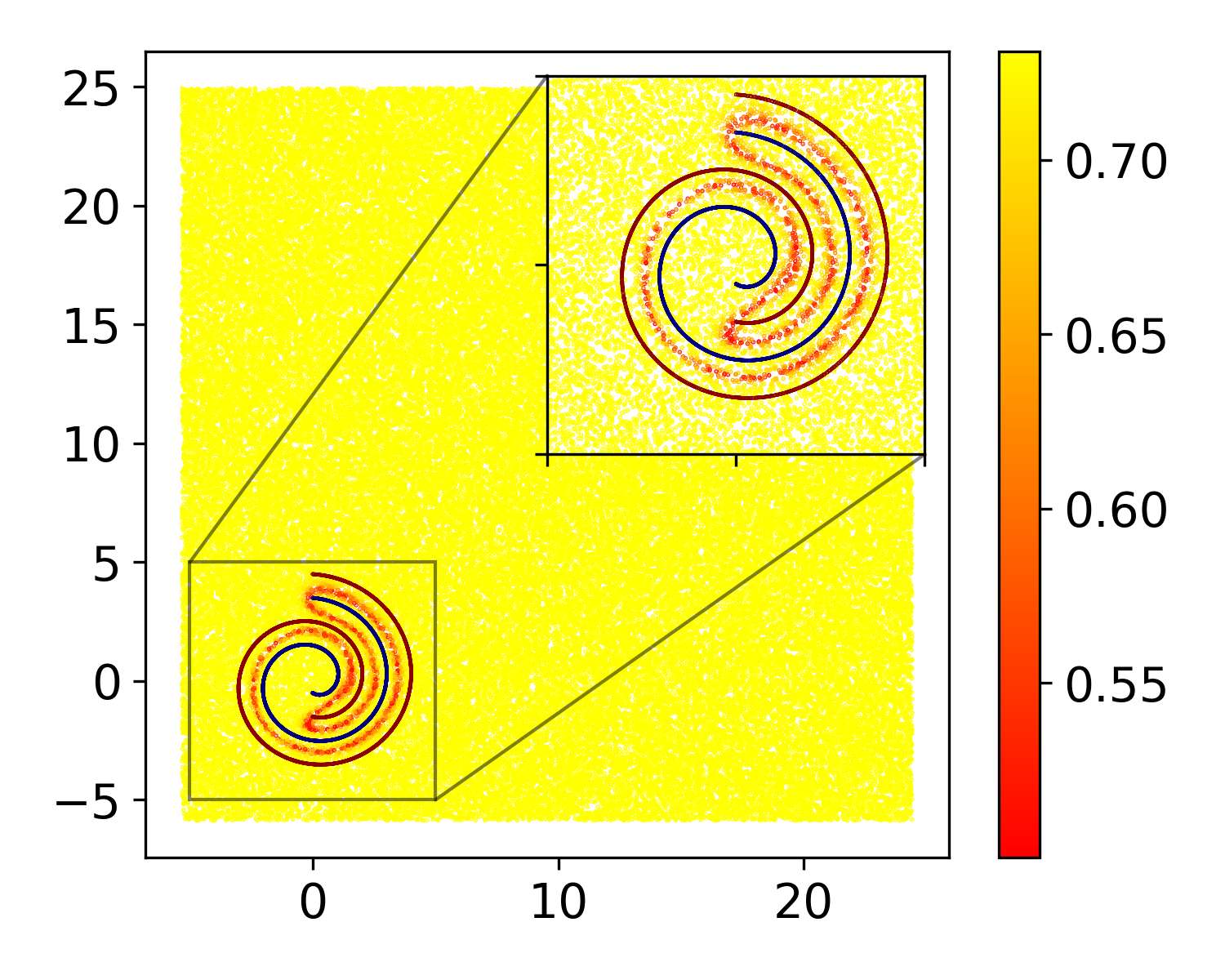}
    \caption{\stdclf{}}
  \end{subfigure} \hspace{.15\linewidth}
  \begin{subfigure}[b]{0.26\linewidth}
    \includegraphics[width=\linewidth]{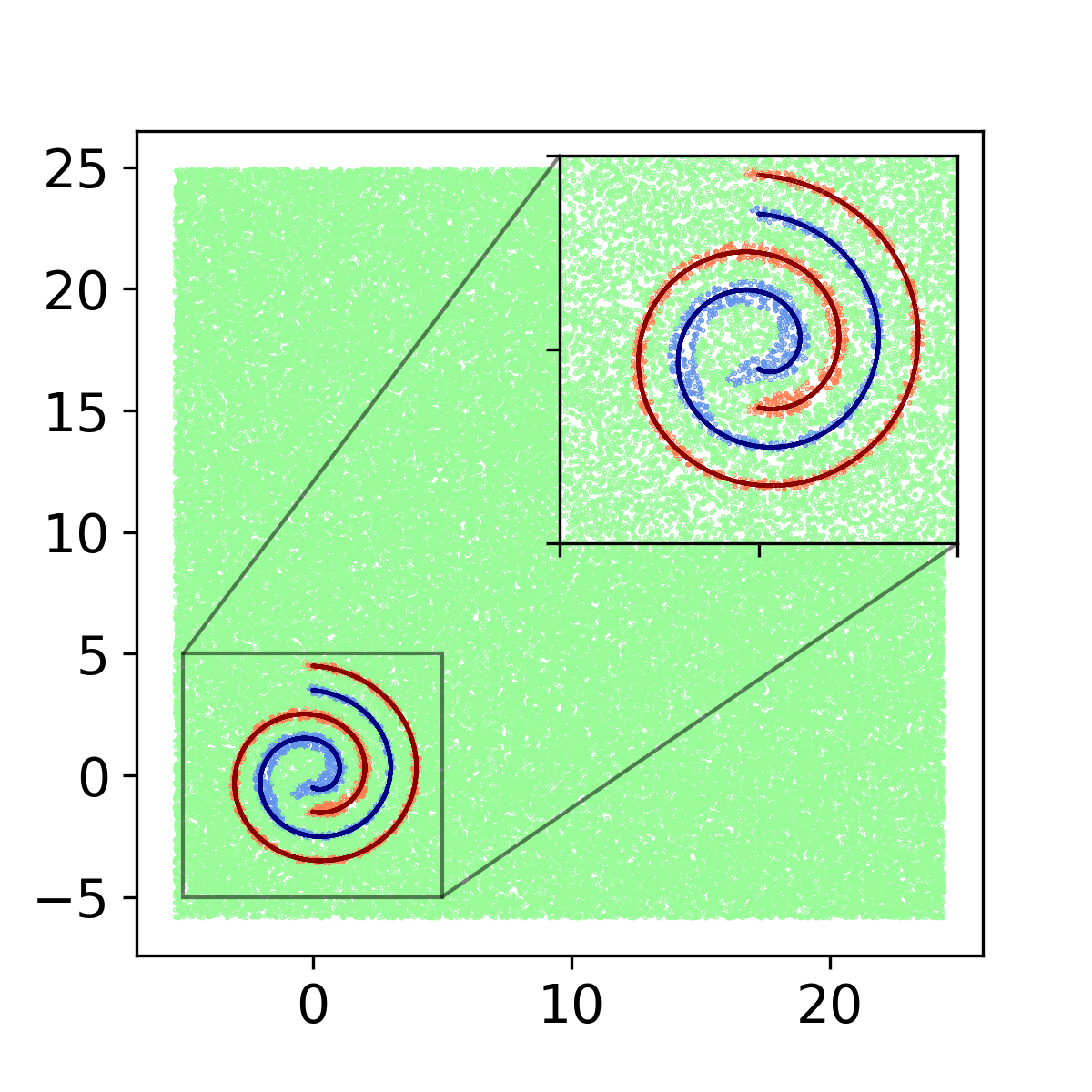}
    \caption{\distlearner{}}
  \end{subfigure}

  \caption{Far region predictions for swiss roll dataset with $m=1, n=500$: (a) Heatmap of prediction confidence for \stdclf{} in a large domain. Even points very far from the manifolds are assigned a class with high confidence of ${\sim}1$. Only for points close to the class manifolds (see inset), meaningful confidence scores are predicted. (b) \distlearner{} decision region in a large domain. Far points are correctly predicted as out-of-domain, while good decision boundaries are obtained in the closer region (inset).}
  \label{fig:confidence_heatmap}
\end{figure}

Unlike \stdclf{}, where all points in the domain are assigned to one of the classes, \distlearner{} only classifies points truly close to a class. This allows \distlearner{} to learn much more meaningful decision boundaries as can be seen in Figure~\ref{fig:decision_boundary}. Another problem with standard classifiers is that, not only do they assign a class to out-of-domain points, they can do so with high confidence~\citep{hein2019relu}. We demonstrate this problem in Figure~\ref{fig:confidence_heatmap}, which shows the confidence (softmax probability) heatmap for points in a 2D slice. As shown, even for points far from any class, \stdclf{} classifies them as one of the classes with high confidence.

\subsection{Adversarial Robustness}
\label{sec:adversarial_robustness}

\begin{figure}
    \centering
    \begin{subfigure}[b]{0.4\linewidth}
    \includegraphics[width=\linewidth]{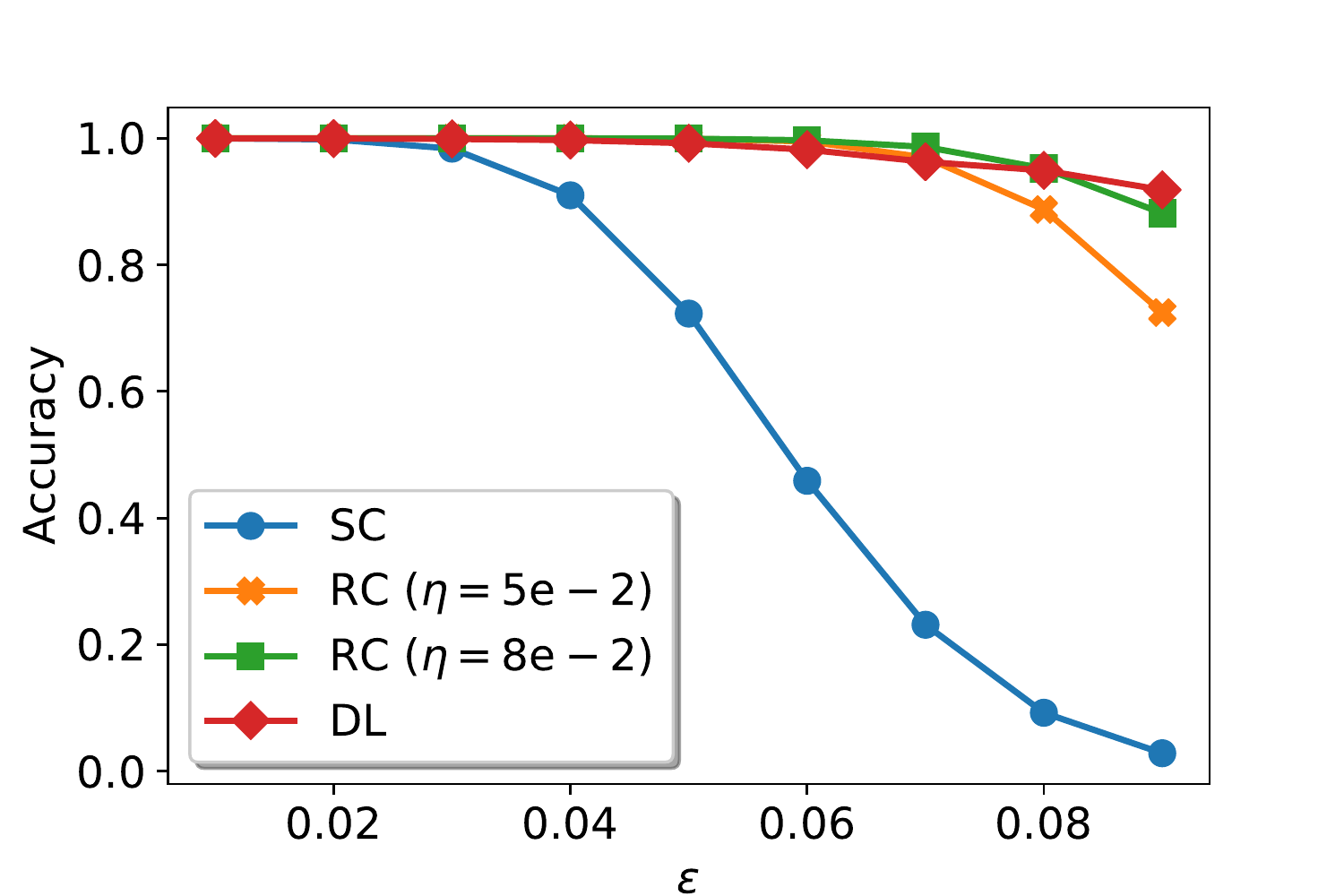}
    \caption{Concentric Spheres}
  \end{subfigure}
  \begin{subfigure}[b]{0.4\linewidth}
    \includegraphics[width=\linewidth]{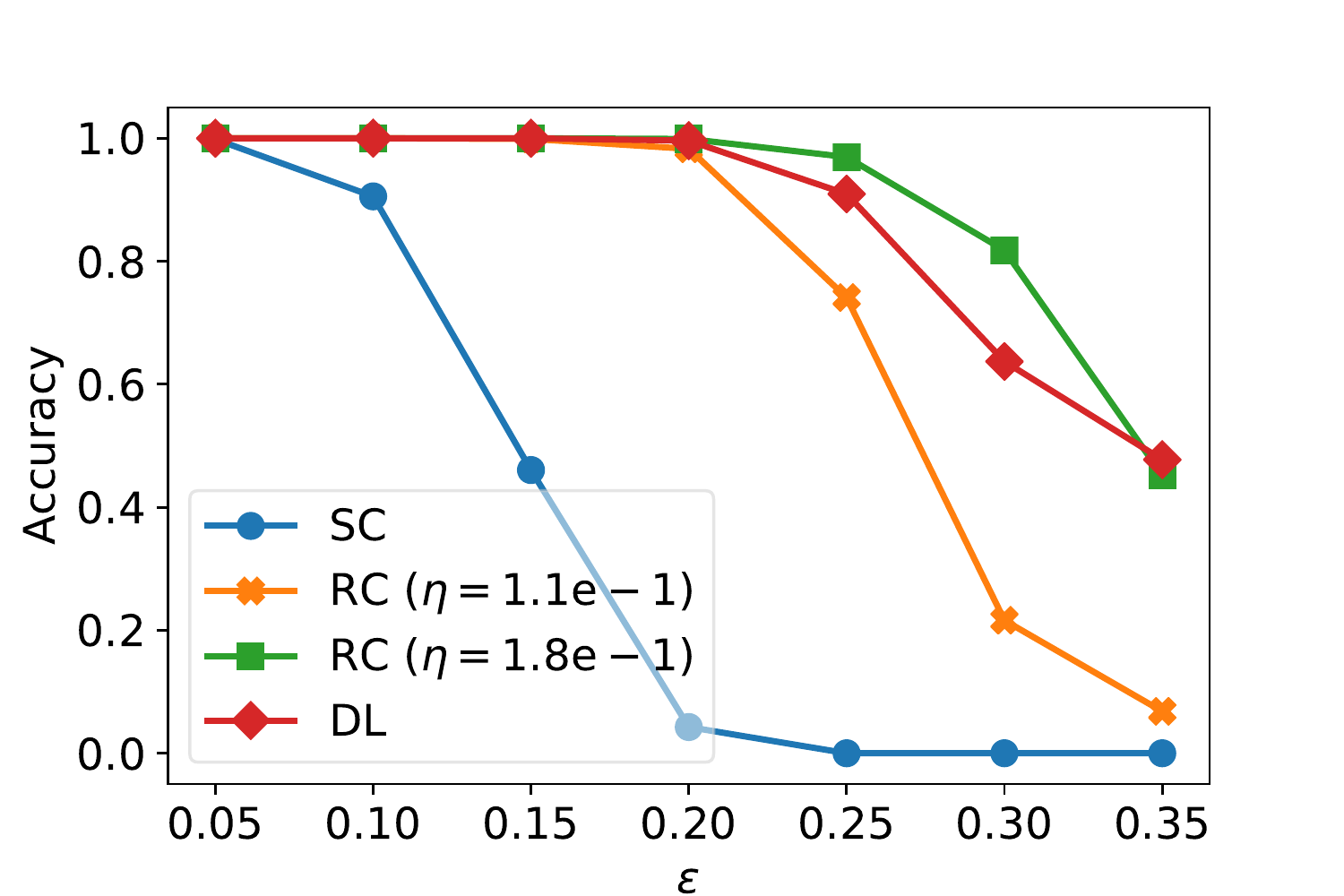}
    \caption{Intertwined Swiss Rolls}
  \end{subfigure}
    \caption{Adversarial Robustness of \distlearner{} (\textbf{DL}), \stdclf{} (\textbf{SC}), and \advclf{} (\textbf{RC}) over different values of $\epsilon$ (maximum adversarial perturbation). \distlearner{} outperforms standard classifier and is comparable to \advclf{}.}
    \label{fig:adversarial_robustness}
\end{figure}

In this section we evaluate the robustness of \distlearner{} to adversarial attacks. For these experiments, we use an additional baseline, \advclf{}, which is the same as \stdclf{} but trained via the adversarial training method (or robust-training) proposed by~\citet{Madry2018TowardsDL}. In robust-training, the model is trained by optimizing the following min-max problem:
\begin{equation}
    \min_\theta \mathop{\mathbb{E}}_{(x,y) \in \mathcal{D}} \left[ \max_{\lVert\delta\rVert_2 < \eta} \mathcal{L}(f(\mathbf{x}_i + \delta;\theta), y_i) \right]
    \label{eq:robust_training}
\end{equation}
where $\theta$ denotes the network parameters, $f$ denotes the network function, and $\eta$ is the search radius for the adversarial perturbation in the inner loop. In all our experiments, the inner loop is run for 40 iterations.

For a sample $(\mathbf{x}, y) \in \mathcal{D}$, we find an adversarial point by maximizing the loss $\mathcal{L}(f(\mathbf{x} + \delta;\theta), y)$ in an $\epsilon$ neighborhood around the point. For \stdclf{} $\mathcal{L}$ corresponds to the cross-entropy loss. For \distlearner{} we maximize the predicted distance to class $y$ minus the predicted distance to the other class, to drive towards misclassification. We solve this constrained optimization problem via the white-box projected gradient descent (PGD) method~\citep{Madry2018TowardsDL}. For each attack we run $100$ PGD iterations with a step size of $\expnumber{5}{-3}$.

We focus on the Concentric Spheres and Intertwined Swiss Rolls datasets for analysis of adversarial robustness. Analysis on Concentric Spheres is inspired by \citet{gilmer2018adversarial}, which used this dataset to extensively analyze the connection between adversarial robustness and high dimensional geometry. For Intertwined Swiss Rolls, we use the analytic form of the normals at each point instead of inferred normal directions for computing off-manifold points, as inferred manifold was not a good approximation of the actual manifold. As discussed in section~\ref{sec:datasets}, we focus on the case of $m=50$ to mimic real image datasets.

Our results are shown in Figure~\ref{fig:adversarial_robustness}. We evaluate adversarial robustness for multiple values of the attack neighborhood radius, $\epsilon$. For \advclf{}, we show results with $\eta \in \{\expnumber{5}{-2}, \expnumber{8}{-2}\}$ for Concentric Spheres and $\eta \in \{\expnumber{1.1}{-1}, \expnumber{1.8}{-1}\}$ for Intertwined Swiss Rolls. We use $tol=0.14$ for Concentric Spheres, while for Intertwined Swiss Rolls, we classify using minimum predicted distance. For higher values of $\eta$, we found the performance of \advclf{} to significantly degrade. As shown in the figure, \distlearner{} not only outperforms the standard classifier significantly, but also performs at par with adversarially trained classifiers. Since \distlearner{} incorporates manifold geometry information into training, we suspect that it learns smoother and more meaningful decision boundaries, allowing it to mitigate vulnerability to adversarial attacks.

\section{Conclusion}
\label{sec:conclusion}
The well known no free lunch theorems drive the point that incorporating prior knowledge about underlying learning problem can significantly improve outcomes. In this paper, we propose a novel method, \distlearner{}, to incorporate the manifold prior while training DNNs. \distlearner{} is trained to predict the distance of a point from the underlying class manifolds. We do this by efficiently generating augmented points at a known distance to the class manifolds. Our evaluation reveals that \distlearner{} is robust to adversarial attacks, and is able to predict distances accurately. Moreover, \distlearner{} is able to identify out-of-domain points and learn much more meaningful decision regions.

\noindent\textbf{Limitations and Future Work:} We acknowledge a few limitations of our current work. Firstly, we have evaluated \distlearner{} only on synthetic datasets. Although it reveals important insights and promise of the current work, we would like to extend our evaluation to real-world datasets. This may require further work on accurately estimating local manifold from training data. Also, in our current method, augmented points are sampled uniformly in small bands around the manifold. However, this may not be the most sample efficient. We want to explore more efficient sampling techniques by incorporating ideas similar to~\citet{goyal2020drocc}. Finally, we would like to extend our method to include unsupervised data. Since inferring the local manifold only relies on nearest neighbors, unsupervised data can be incorporated for this step. Given the abundant availability of unsupervised data, this can provide significant gains.

\bibliographystyle{unsrtnat}
\bibliography{bibliography}

\begin{thebibliography}{42}
\providecommand{\natexlab}[1]{#1}
\providecommand{\url}[1]{\texttt{#1}}
\expandafter\ifx\csname urlstyle\endcsname\relax
  \providecommand{\doi}[1]{doi: #1}\else
  \providecommand{\doi}{doi: \begingroup \urlstyle{rm}\Url}\fi

\bibitem[Domingos(2012)]{pedro_few_useful_things}
Pedro Domingos.
\newblock A few useful things to know about machine learning.
\newblock \emph{Commun. ACM}, 55\penalty0 (10):\penalty0 78–87, oct 2012.
\newblock ISSN 0001-0782.
\newblock \doi{10.1145/2347736.2347755}.
\newblock URL \url{https://doi.org/10.1145/2347736.2347755}.

\bibitem[Yu et~al.(2022)Yu, Wang, Vasudevan, Yeung, Seyedhosseini, and
  Wu]{cocaimgclf}
Jiahui Yu, Zirui Wang, Vijay Vasudevan, Legg Yeung, Mojtaba Seyedhosseini, and
  Yonghui Wu.
\newblock Coca: Contrastive captioners are image-text foundation models, 2022.
\newblock URL \url{https://arxiv.org/abs/2205.01917}.

\bibitem[Devlin et~al.(2019)Devlin, Chang, Lee, and
  Toutanova]{devlin-etal-2019-bert}
Jacob Devlin, Ming-Wei Chang, Kenton Lee, and Kristina Toutanova.
\newblock {BERT}: Pre-training of deep bidirectional transformers for language
  understanding.
\newblock In \emph{Proceedings of the 2019 Conference of the North {A}merican
  Chapter of the Association for Computational Linguistics: Human Language
  Technologies, Volume 1 (Long and Short Papers)}, pages 4171--4186,
  Minneapolis, Minnesota, June 2019. Association for Computational Linguistics.
\newblock \doi{10.18653/v1/N19-1423}.
\newblock URL \url{https://aclanthology.org/N19-1423}.

\bibitem[Raffel et~al.(2020)Raffel, Shazeer, Roberts, Lee, Narang, Matena,
  Zhou, Li, and Liu]{2020t5}
Colin Raffel, Noam Shazeer, Adam Roberts, Katherine Lee, Sharan Narang, Michael
  Matena, Yanqi Zhou, Wei Li, and Peter~J. Liu.
\newblock Exploring the limits of transfer learning with a unified text-to-text
  transformer.
\newblock \emph{Journal of Machine Learning Research}, 21\penalty0
  (140):\penalty0 1--67, 2020.
\newblock URL \url{http://jmlr.org/papers/v21/20-074.html}.

\bibitem[Shen et~al.(2018)Shen, Pang, Weiss, Schuster, Jaitly, Yang, Chen,
  Zhang, Wang, Skerry-Ryan, Saurous, Agiomyrgiannakis, and
  Wu]{Shen2018NaturalTS}
Jonathan Shen, Ruoming Pang, Ron~J. Weiss, Mike Schuster, Navdeep Jaitly,
  Zongheng Yang, Z.~Chen, Yu~Zhang, Yuxuan Wang, R.~J. Skerry-Ryan, Rif~A.
  Saurous, Yannis Agiomyrgiannakis, and Yonghui Wu.
\newblock Natural tts synthesis by conditioning wavenet on mel spectrogram
  predictions.
\newblock \emph{2018 IEEE International Conference on Acoustics, Speech and
  Signal Processing (ICASSP)}, pages 4779--4783, 2018.

\bibitem[Wolpert and Macready(1997)]{wolpert1997no_free_lunch_optimization}
David~H Wolpert and William~G Macready.
\newblock No free lunch theorems for optimization.
\newblock \emph{IEEE transactions on evolutionary computation}, 1\penalty0
  (1):\penalty0 67--82, 1997.

\bibitem[Wolpert(1996)]{wolpert1996lack_of_apriori}
David~H Wolpert.
\newblock The lack of a priori distinctions between learning algorithms.
\newblock \emph{Neural computation}, 8\penalty0 (7):\penalty0 1341--1390, 1996.

\bibitem[Wolpert(2002)]{wolpert2002supervised_no_free_lunch}
David~H Wolpert.
\newblock The supervised learning no-free-lunch theorems.
\newblock \emph{Soft computing and industry}, pages 25--42, 2002.

\bibitem[Vaswani et~al.(2017)Vaswani, Shazeer, Parmar, Uszkoreit, Jones, Gomez,
  Kaiser, and Polosukhin]{vaswani_attention}
Ashish Vaswani, Noam Shazeer, Niki Parmar, Jakob Uszkoreit, Llion Jones,
  Aidan~N Gomez, \L~ukasz Kaiser, and Illia Polosukhin.
\newblock Attention is all you need.
\newblock In I.~Guyon, U.~Von Luxburg, S.~Bengio, H.~Wallach, R.~Fergus,
  S.~Vishwanathan, and R.~Garnett, editors, \emph{Advances in Neural
  Information Processing Systems}, volume~30. Curran Associates, Inc., 2017.
\newblock URL
  \url{https://proceedings.neurips.cc/paper/2017/file/3f5ee243547dee91fbd053c1c4a845aa-Paper.pdf}.

\bibitem[Cohen and Welling(2016)]{cohen_equivariant_cnns}
Taco Cohen and Max Welling.
\newblock Group equivariant convolutional networks.
\newblock In Maria~Florina Balcan and Kilian~Q. Weinberger, editors,
  \emph{Proceedings of The 33rd International Conference on Machine Learning},
  volume~48 of \emph{Proceedings of Machine Learning Research}, pages
  2990--2999, New York, New York, USA, 20--22 Jun 2016. PMLR.
\newblock URL \url{https://proceedings.mlr.press/v48/cohenc16.html}.

\bibitem[Bronstein et~al.(2021)Bronstein, Bruna, Cohen, and
  Velickovic]{gdl_book}
Michael~M. Bronstein, Joan Bruna, Taco Cohen, and Petar Velickovic.
\newblock Geometric deep learning: Grids, groups, graphs, geodesics, and
  gauges.
\newblock \emph{CoRR}, abs/2104.13478, 2021.
\newblock URL \url{https://arxiv.org/abs/2104.13478}.

\bibitem[Bietti et~al.(2021)Bietti, Venturi, and Bruna]{gdl_sample_complexity}
Alberto Bietti, Luca Venturi, and Joan Bruna.
\newblock On the sample complexity of learning under geometric stability.
\newblock In M.~Ranzato, A.~Beygelzimer, Y.~Dauphin, P.S. Liang, and J.~Wortman
  Vaughan, editors, \emph{Advances in Neural Information Processing Systems},
  volume~34, pages 18673--18684. Curran Associates, Inc., 2021.
\newblock URL
  \url{https://proceedings.neurips.cc/paper/2021/file/9ac5a6d86e8924182271bd820acbce0e-Paper.pdf}.

\bibitem[Tenenbaum et~al.(2000)Tenenbaum, Silva, and
  Langford]{tenenbaum2000isomap_nonlinear_dim_reduction}
Joshua~B Tenenbaum, Vin~de Silva, and John~C Langford.
\newblock A global geometric framework for nonlinear dimensionality reduction.
\newblock \emph{science}, 290\penalty0 (5500):\penalty0 2319--2323, 2000.

\bibitem[Roweis and Saul(2000)]{roweis2000nonlinear_LLE}
Sam~T Roweis and Lawrence~K Saul.
\newblock Nonlinear dimensionality reduction by locally linear embedding.
\newblock \emph{science}, 290\penalty0 (5500):\penalty0 2323--2326, 2000.

\bibitem[Zhang and Zha(2004{\natexlab{a}})]{zhang2004tangent_space_alignment}
Zhenyue Zhang and Hongyuan Zha.
\newblock Principal manifolds and nonlinear dimensionality reduction via
  tangent space alignment.
\newblock \emph{SIAM journal on scientific computing}, 26\penalty0
  (1):\penalty0 313--338, 2004{\natexlab{a}}.

\bibitem[Rifai et~al.(2011)Rifai, Dauphin, Vincent, Bengio, and
  Muller]{rifai2011manifold_tangent_classifier}
Salah Rifai, Yann~N Dauphin, Pascal Vincent, Yoshua Bengio, and Xavier Muller.
\newblock The manifold tangent classifier.
\newblock \emph{Advances in neural information processing systems}, 24, 2011.

\bibitem[Hein et~al.(2019)Hein, Andriushchenko, and Bitterwolf]{hein2019relu}
Matthias Hein, Maksym Andriushchenko, and Julian Bitterwolf.
\newblock Why relu networks yield high-confidence predictions far away from the
  training data and how to mitigate the problem.
\newblock In \emph{Proceedings of the IEEE/CVF Conference on Computer Vision
  and Pattern Recognition}, pages 41--50, 2019.

\bibitem[Goodfellow et~al.(2014)Goodfellow, Shlens, and
  Szegedy]{goodfellow2014explaining}
Ian~J Goodfellow, Jonathon Shlens, and Christian Szegedy.
\newblock Explaining and harnessing adversarial examples.
\newblock \emph{arXiv preprint arXiv:1412.6572}, 2014.

\bibitem[Huang et~al.(2017)Huang, Papernot, Goodfellow, Duan, and
  Abbeel]{huang2017adversarial}
Sandy Huang, Nicolas Papernot, Ian Goodfellow, Yan Duan, and Pieter Abbeel.
\newblock Adversarial attacks on neural network policies.
\newblock \emph{arXiv preprint arXiv:1702.02284}, 2017.

\bibitem[Akhtar and Mian(2018)]{akhtar2018threat_adversarial}
Naveed Akhtar and Ajmal Mian.
\newblock Threat of adversarial attacks on deep learning in computer vision: A
  survey.
\newblock \emph{Ieee Access}, 6:\penalty0 14410--14430, 2018.

\bibitem[Chakraborty et~al.(2018)Chakraborty, Alam, Dey, Chattopadhyay, and
  Mukhopadhyay]{chakraborty2018adversarial}
Anirban Chakraborty, Manaar Alam, Vishal Dey, Anupam Chattopadhyay, and Debdeep
  Mukhopadhyay.
\newblock Adversarial attacks and defences: A survey.
\newblock \emph{arXiv preprint arXiv:1810.00069}, 2018.

\bibitem[Madry et~al.(2018)Madry, Makelov, Schmidt, Tsipras, and
  Vladu]{Madry2018TowardsDL}
Aleksander Madry, Aleksandar Makelov, Ludwig Schmidt, Dimitris Tsipras, and
  Adrian Vladu.
\newblock Towards deep learning models resistant to adversarial attacks.
\newblock \emph{ArXiv}, abs/1706.06083, 2018.

\bibitem[Gilmer et~al.(2018)Gilmer, Metz, Faghri, Schoenholz, Raghu,
  Wattenberg, and Goodfellow]{gilmer2018adversarial}
Justin Gilmer, Luke Metz, Fartash Faghri, Sam Schoenholz, Maithra Raghu, Martin
  Wattenberg, and Ian Goodfellow.
\newblock Adversarial spheres, 2018.
\newblock URL \url{https://openreview.net/forum?id=SyUkxxZ0b}.

\bibitem[Stutz et~al.(2019)Stutz, Hein, and Schiele]{Stutz2019CVPR}
David Stutz, Matthias Hein, and Bernt Schiele.
\newblock Disentangling adversarial robustness and generalization.
\newblock 2019.

\bibitem[Jolliffe(1986)]{Jolliffe:1986:pca}
I.T. Jolliffe.
\newblock \emph{Principal Component Analysis}.
\newblock Springer Verlag, 1986.

\bibitem[Zhang and Zha(2004{\natexlab{b}})]{LTSA}
Zhenyue Zhang and Hongyuan Zha.
\newblock Principal manifolds and nonlinear dimensionality reduction via
  tangent space alignment.
\newblock \emph{SIAM Journal on Scientific Computing}, 26\penalty0
  (1):\penalty0 313--338, 2004{\natexlab{b}}.
\newblock \doi{10.1137/S1064827502419154}.
\newblock URL \url{https://doi.org/10.1137/S1064827502419154}.

\bibitem[Lin and Zha(2008)]{lin2008riemannian}
Tong Lin and Hongbin Zha.
\newblock Riemannian manifold learning.
\newblock \emph{IEEE Transactions on Pattern Analysis and Machine
  Intelligence}, 30\penalty0 (5):\penalty0 796--809, 2008.

\bibitem[Osher and Sethian(1988)]{osher1988fronts}
Stanley Osher and James~A Sethian.
\newblock Fronts propagating with curvature-dependent speed: Algorithms based
  on hamilton-jacobi formulations.
\newblock \emph{Journal of computational physics}, 79\penalty0 (1):\penalty0
  12--49, 1988.

\bibitem[Osher and Fedkiw(2001)]{osher2001level}
Stanley Osher and Ronald~P Fedkiw.
\newblock Level set methods: an overview and some recent results.
\newblock \emph{Journal of Computational physics}, 169\penalty0 (2):\penalty0
  463--502, 2001.

\bibitem[Enright et~al.(2002)Enright, Fedkiw, Ferziger, and
  Mitchell]{enright2002hybrid}
Douglas Enright, Ronald Fedkiw, Joel Ferziger, and Ian Mitchell.
\newblock A hybrid particle level set method for improved interface capturing.
\newblock \emph{Journal of Computational physics}, 183\penalty0 (1):\penalty0
  83--116, 2002.

\bibitem[Park et~al.(2019)Park, Florence, Straub, Newcombe, and
  Lovegrove]{deepsdf}
Jeong~Joon Park, Peter Florence, Julian Straub, Richard~A. Newcombe, and Steven
  Lovegrove.
\newblock Deepsdf: Learning continuous signed distance functions for shape
  representation.
\newblock In \emph{{IEEE} Conference on Computer Vision and Pattern
  Recognition, {CVPR} 2019, Long Beach, CA, USA, June 16-20, 2019}, pages
  165--174. Computer Vision Foundation / {IEEE}, 2019.
\newblock \doi{10.1109/CVPR.2019.00025}.
\newblock URL
  \url{http://openaccess.thecvf.com/content\_CVPR\_2019/html/Park\_DeepSDF\_Learning\_Continuous\_Signed\_Distance\_Functions\_for\_Shape\_Representation\_CVPR\_2019\_paper.html}.

\bibitem[Kingma and Ba(2015)]{adam}
Diederik~P. Kingma and Jimmy Ba.
\newblock Adam: {A} method for stochastic optimization.
\newblock In Yoshua Bengio and Yann LeCun, editors, \emph{3rd International
  Conference on Learning Representations, {ICLR} 2015, San Diego, CA, USA, May
  7-9, 2015, Conference Track Proceedings}, 2015.
\newblock URL \url{http://arxiv.org/abs/1412.6980}.

\bibitem[Marsland(2009)]{swiss_roll_dataset_original}
Stephen Marsland.
\newblock \emph{Machine Learning - An Algorithmic Perspective.}
\newblock Chapman and Hall / CRC machine learning and pattern recognition
  series. CRC Press, 2009.
\newblock ISBN 978-1-4200-6718-7.

\bibitem[Pope et~al.(2021)Pope, Zhu, Abdelkader, Goldblum, and
  Goldstein]{intrinsic_dimension}
Phil Pope, Chen Zhu, Ahmed Abdelkader, Micah Goldblum, and Tom Goldstein.
\newblock The intrinsic dimension of images and its impact on learning.
\newblock In \emph{International Conference on Learning Representations}, 2021.
\newblock URL \url{https://openreview.net/forum?id=XJk19XzGq2J}.

\bibitem[Goyal et~al.(2020)Goyal, Raghunathan, Jain, Simhadri, and
  Jain]{goyal2020drocc}
Sachin Goyal, Aditi Raghunathan, Moksh Jain, Harsha~Vardhan Simhadri, and
  Prateek Jain.
\newblock Drocc: Deep robust one-class classification.
\newblock In \emph{ICML 2020}, July 2020.
\newblock URL
  \url{https://www.microsoft.com/en-us/research/publication/drocc-deep-robust-one-class-classification/}.

\bibitem[Ioffe and Szegedy(2015)]{batch_norm}
Sergey Ioffe and Christian Szegedy.
\newblock Batch normalization: Accelerating deep network training by reducing
  internal covariate shift.
\newblock In Francis Bach and David Blei, editors, \emph{Proceedings of the
  32nd International Conference on Machine Learning}, volume~37 of
  \emph{Proceedings of Machine Learning Research}, pages 448--456, Lille,
  France, 07--09 Jul 2015. PMLR.
\newblock URL \url{https://proceedings.mlr.press/v37/ioffe15.html}.

\bibitem[Ruder(2017)]{ruder_mtl_survey}
Sebastian Ruder.
\newblock An overview of multi-task learning in deep neural networks.
\newblock \emph{CoRR}, abs/1706.05098, 2017.
\newblock URL \url{http://arxiv.org/abs/1706.05098}.

\bibitem[Paszke et~al.(2019)Paszke, Gross, Massa, Lerer, Bradbury, Chanan,
  Killeen, Lin, Gimelshein, Antiga, Desmaison, Kopf, Yang, DeVito, Raison,
  Tejani, Chilamkurthy, Steiner, Fang, Bai, and Chintala]{pytorch}
Adam Paszke, Sam Gross, Francisco Massa, Adam Lerer, James Bradbury, Gregory
  Chanan, Trevor Killeen, Zeming Lin, Natalia Gimelshein, Luca Antiga, Alban
  Desmaison, Andreas Kopf, Edward Yang, Zachary DeVito, Martin Raison, Alykhan
  Tejani, Sasank Chilamkurthy, Benoit Steiner, Lu~Fang, Junjie Bai, and Soumith
  Chintala.
\newblock Pytorch: An imperative style, high-performance deep learning library.
\newblock In H.~Wallach, H.~Larochelle, A.~Beygelzimer, F.~d\textquotesingle
  Alch\'{e}-Buc, E.~Fox, and R.~Garnett, editors, \emph{Advances in Neural
  Information Processing Systems 32}, pages 8024--8035. Curran Associates,
  Inc., 2019.
\newblock URL
  \url{http://papers.neurips.cc/paper/9015-pytorch-an-imperative-style-high-performance-deep-learning-library.pdf}.

\bibitem[Harris et~al.(2020)Harris, Millman, van~der Walt, Gommers, Virtanen,
  Cournapeau, Wieser, Taylor, Berg, Smith, Kern, Picus, Hoyer, van Kerkwijk,
  Brett, Haldane, del R{\'{i}}o, Wiebe, Peterson, G{\'{e}}rard-Marchant,
  Sheppard, Reddy, Weckesser, Abbasi, Gohlke, and Oliphant]{numpy}
Charles~R. Harris, K.~Jarrod Millman, St{\'{e}}fan~J. van~der Walt, Ralf
  Gommers, Pauli Virtanen, David Cournapeau, Eric Wieser, Julian Taylor,
  Sebastian Berg, Nathaniel~J. Smith, Robert Kern, Matti Picus, Stephan Hoyer,
  Marten~H. van Kerkwijk, Matthew Brett, Allan Haldane, Jaime~Fern{\'{a}}ndez
  del R{\'{i}}o, Mark Wiebe, Pearu Peterson, Pierre G{\'{e}}rard-Marchant,
  Kevin Sheppard, Tyler Reddy, Warren Weckesser, Hameer Abbasi, Christoph
  Gohlke, and Travis~E. Oliphant.
\newblock Array programming with {NumPy}.
\newblock \emph{Nature}, 585\penalty0 (7825):\penalty0 357--362, September
  2020.
\newblock \doi{10.1038/s41586-020-2649-2}.
\newblock URL \url{https://doi.org/10.1038/s41586-020-2649-2}.

\bibitem[Virtanen et~al.(2020)Virtanen, Gommers, Oliphant, Haberland, Reddy,
  Cournapeau, Burovski, Peterson, Weckesser, Bright, {van der Walt}, Brett,
  Wilson, Millman, Mayorov, Nelson, Jones, Kern, Larson, Carey, Polat, Feng,
  Moore, {VanderPlas}, Laxalde, Perktold, Cimrman, Henriksen, Quintero, Harris,
  Archibald, Ribeiro, Pedregosa, {van Mulbregt}, and {SciPy 1.0
  Contributors}]{scipy}
Pauli Virtanen, Ralf Gommers, Travis~E. Oliphant, Matt Haberland, Tyler Reddy,
  David Cournapeau, Evgeni Burovski, Pearu Peterson, Warren Weckesser, Jonathan
  Bright, St{\'e}fan~J. {van der Walt}, Matthew Brett, Joshua Wilson, K.~Jarrod
  Millman, Nikolay Mayorov, Andrew R.~J. Nelson, Eric Jones, Robert Kern, Eric
  Larson, C~J Carey, {\.I}lhan Polat, Yu~Feng, Eric~W. Moore, Jake
  {VanderPlas}, Denis Laxalde, Josef Perktold, Robert Cimrman, Ian Henriksen,
  E.~A. Quintero, Charles~R. Harris, Anne~M. Archibald, Ant{\^o}nio~H. Ribeiro,
  Fabian Pedregosa, Paul {van Mulbregt}, and {SciPy 1.0 Contributors}.
\newblock {{SciPy} 1.0: Fundamental Algorithms for Scientific Computing in
  Python}.
\newblock \emph{Nature Methods}, 17:\penalty0 261--272, 2020.
\newblock \doi{10.1038/s41592-019-0686-2}.

\bibitem[Pedregosa et~al.(2011)Pedregosa, Varoquaux, Gramfort, Michel, Thirion,
  Grisel, Blondel, Prettenhofer, Weiss, Dubourg, Vanderplas, Passos,
  Cournapeau, Brucher, Perrot, and Duchesnay]{scikit-learn}
F.~Pedregosa, G.~Varoquaux, A.~Gramfort, V.~Michel, B.~Thirion, O.~Grisel,
  M.~Blondel, P.~Prettenhofer, R.~Weiss, V.~Dubourg, J.~Vanderplas, A.~Passos,
  D.~Cournapeau, M.~Brucher, M.~Perrot, and E.~Duchesnay.
\newblock Scikit-learn: Machine learning in {P}ython.
\newblock \emph{Journal of Machine Learning Research}, 12:\penalty0 2825--2830,
  2011.

\bibitem[Iyer et~al.(2020)Iyer, Thejas, Kwatra, Ramjee, and
  Sivathanu]{iyer2020wide}
Nikhil Iyer, V~Thejas, Nipun Kwatra, Ramachandran Ramjee, and Muthian
  Sivathanu.
\newblock Wide-minima density hypothesis and the explore-exploit learning rate
  schedule.
\newblock \emph{arXiv preprint arXiv:2003.03977}, 2020.

\end{thebibliography}

\newpage
\appendix
\section{Model Architecture}

\begin{figure}[!h]
    \centering
    \includegraphics[width=\linewidth]{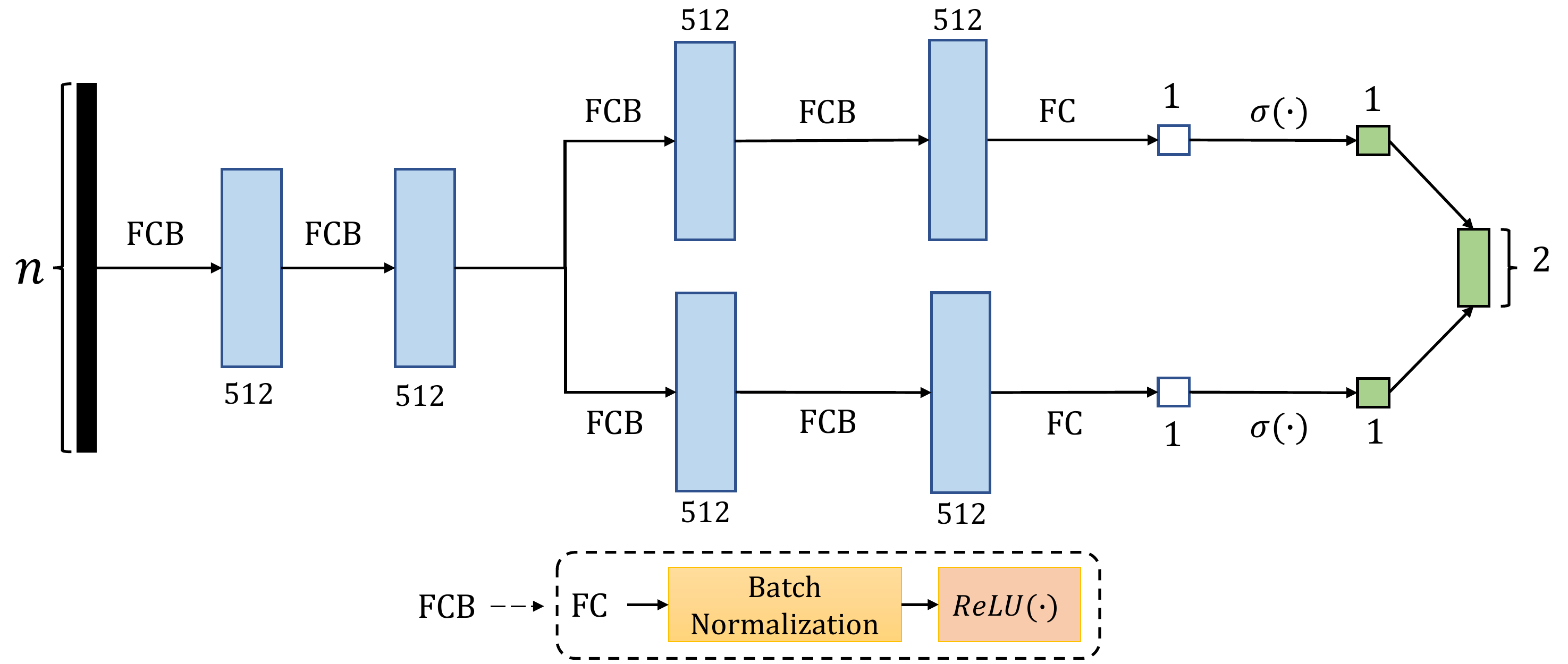}
    \caption{\distlearner{} model architecture. The boxes denote tensors and arrows denote operations. $\mathbf{FC}$ refers to a fully-connected layer, and $\mathbf{FCB}$ refers to a fully-connected layer followed by batch normalization and $\text{ReLU}$ activation.}
    \label{fig:model}
\end{figure}

Figure~\ref{fig:model} shows the architecture of our model. The first two layers of our model consist of $\mathbf{FCB}$ blocks (fully connected layers of 512 neurons, followed by batch normalization \citep{batch_norm} and a $\text{ReLU}$ activation). After this, the network branches into multiple pathways, one for each manifold. Each branch consists of two more $\mathbf{FCB}$ blocks, followed by a fully connected layer ($\mathbf{FC}$) and finally a sigmoid activation ($\sigma$). The output of each branch gives the distance of the point from the branch's corresponding manifold. We collect the outputs of each branch into a single tensor which is then used for computing MSE loss. The branched network architecture was inspired by hard parameter sharing techniques used in multi-task learning~\citep{ruder_mtl_survey}. In our experiments, we found the branched architecture to improve the quality of the learnt distance function.

We used PyTorch ~\citep{pytorch} framework to build our model. Other tools used included Numpy ~\citep{numpy}, Scipy ~\citep{scipy}, and Scikit-learn ~\citep{scikit-learn} (all released under BSD license). All our experiments were performed on a V100 GPU with 16GB memory. Our code is available under MIT license at: \url{https://github.com/microsoft/distance-learner}

\section{Synthetic Data Generation}
\label{sec:appendix-data-gen}
To generate our synthetic datasets, we need to sample points on $m$-dimensional manifolds, embedded in $n$-dimensional space ($\mathbb{R}^n$). Our sampling method is broken into three steps -- 1) sampling the $m$-dimensional manifold in an $m+1$-dimensional embedding (we call this canonical embedding), 2) embedding these points in $\mathbb{R}^n$ via what we call \textit{trivial} embedding and 3) applying a random transformation.%

\subsection{Sampling Canonical Embeddings}

We first sample points on the $m$-dimensional manifold embedded in $\mathbb{R}^{m+1}$ (we call this the canonical embedding). This process depends on the dataset that we are using. 

\noindent\textbf{Separated Spheres:} Sampling the canonical embeddings amounts to sampling points uniformly from an $m$-sphere of required radius $r$ and centre $\mathbf{c}_m$ in $\mathbb{R}^{m+1}$. As a first step, we sample points uniformly from a unit-sphere centred at the origin using the steps described in  ~\cite{gilmer2018adversarial}, i.e., sampling a point $\mathbf{z} \in \mathbb{R}^{m + 1}$ from the standard normal distribution, $\mathcal{N}(\vec{0}, I)$, and the scaling $\mathbf{z}$, as follows: $\mathbf{y} = r \frac{\mathbf{z}}{\lVert \mathbf{z} \rVert}$. Finally, to centre the sphere at $\mathbf{c}_m$ apply a translation, $\mathbf{x} = \mathbf{y} + \mathbf{c}_m$. We generate centres by first sampling a random centre for one of the spheres, and then perturbing it using a random perturbation of size $d_c$ for the other sphere. We choose $d_c$ so that there is no overlap between the two spheres. 

\noindent\textbf{Intertwined Swiss Rolls:} We discussed how the Swiss Roll is parameterized in the main text (see section \ref{sec:datasets}).
In order to obtain intertwined Swiss Rolls, we generate the inner Swiss Roll by changing the parameterization to $\mathbf{x}(\phi, \psi_1, \psi_2, \ldots \psi_{m - 1}) = ((\phi - \mu) \cos \phi, (\phi - \mu) \sin \phi, \psi_1, \psi_2, \ldots \psi_3)$, where $\mu$ controls the gap between the Swiss Rolls, and is chosen appropriately to prevent overlap. For our experiments, we use $\phi \in [1.5, 4.5]$, $\psi \in [0, 21]$, and $\mu = 1$.

\noindent\textbf{Concentric Spheres:} We follow the same process as Separated Spheres, except that only one common centre is sampled for both the spheres, and the difference between the radii of the two spheres is chosen suitably to prevent overlap. 

\subsection{Generating Trivial Embeddings}
\label{sec:appendix-trivial-embed}
Let the canonical embeddings of the samples be stored as rows of a matrix $\mathbf{P}_{c}$ with $(m+1)$ columns. Once $\mathbf{P}_{c}$ is obtained, we embed the points trivially in $\mathbb{R}^n$ by concatenating the points with $(n - m - 1)$ 0's. That is, the trivial embedding $\mathbf{P}_{\text{tr}}$ can be obtained as $\mathbf{P}_{\text{tr}} = \begin{bmatrix}\mathbf{P}_{\text{c}}\ & \mathbf{0}\end{bmatrix}$, where $\mathbf{0}$ is a zero matrix.

\subsection{Random Transformations}
\label{sec:appendix-random-tr}
Note that the embedding of the manifold in $\mathbb{R}^n$ is still a trivial one, obtained by concatenating the canonical embeddings with 0's. To make this embedding more generalized, we apply random translation and rotation transforms to the data. In order to generate a random rotation transform, $\mathbf{Q} \in \mathbb{R}^{n \times n}$, we sample a random matrix $\mathbf{M} \in \mathbb{R}^{n \times n}$, and decompose it using QR factorization to obtain an orthogonal matrix $\mathbf{Q}$ and a residual matrix $\mathbf{R}$. The orthogonal matrix $\mathbf{Q}$ can be used as the rotation transform. We sample a random vector, $\vec{\mathbf{T}} \in \mathbb{R}^n$ as a translation transform. We obtain the final transformed samples matrix $\mathbf{P}$ by applying these transforms row-wise to $\mathbf{P}_{\text{tr}}$. For training, we normalize these points so that they lie in a unit-hypercube. This helped significantly with training stability.

\section{Hyperparameters Used}
\label{sec:appendix_hparams}
In this section we describe the various hyperparameter settings used in our experiments. Table~\ref{tab:common_hparams} shows the hyperparameter values used for training. For all our experiments, we used the \textit{Knee} learning rate schedule~\citep{iyer2020wide}, and increase the learning rate linearly from zero to its maximum value till epoch 10, and then decrease linearly to zero from epoch 700 to 1000. All models were trained for 1000 epochs.

\begin{table}[!ht]
    \small
    \centering
    \caption{Values of hyperparameters for \distlearner{}. $\star$ denotes that a hyperparameter sweep was conducted for this dataset. }
    \begin{tabular}{p{1.5cm}cc|cp{1.0cm}p{1.2cm}ccc}
    \toprule
        Dataset & $m$ & $n$ & $max\_norm$ & $N_{\text{on}}$ \tiny{($\times 10^6$)} & $N_{\text{off}}$ \tiny{($\times 10^6$)} & Learning Rate & Batch Size &   \\ \midrule
        \multirow{4}{2cm}{Separated Spheres} & 1 & 2 & 0.10 & 0.50 & 1.00 & $1.0 \times 10^{-5}$ & 512 & ~  \\ 
         & 1 & 50 & 0.10 & 0.50 & 1.00 & $1.0 \times 10^{-5}$ & 512 & ~  \\ 
         & 1 & 500 & 0.10 & 0.50 & 1.00 & $1.5 \times 10^{-5}$ & 4096 & $\star$ \\ 
         & 2 & 500 & 0.10 & 0.05 & 1.00 & $1.0 \times 10^{-5}$ & 512 & ~  \\ \midrule
        \multirow{3}{2cm}{Intertwined Swiss Rolls} & 1 & 2 & 0.40 & 0.05 & 0.05 & $1.0 \times 10^{-5}$ & 512 & $\star$ \\ 
         & 1 & 50 & 0.40 & 0.05 & 0.05 & $1.0 \times 10^{-5}$ & 512 & ~ \\ 
         & 1 & 500 & 0.40 & 0.05 & 1.00 & $1.0 \times 10^{-6}$ & 4096 & $\star$ \\ \midrule
        \multirow{5}{2cm}{Concentric Spheres} & 1 & 2 & 0.10 & 0.50 & 1.00 & $1.5 \times 10^{-5}$ & 4096 & ~  \\ 
         & 1 & 50 & 0.10 & 0.50 & 1.00 & $1.5 \times 10^{-5}$ & 4096 & ~ \\
         & 2 & 50 & 0.14 & 0.50 & 2.00 & $1.5 \times 10^{-5}$ & 4096 & ~ \\
         & 25 & 500 & 0.14 & 0.50 & 6.00 & $1.5 \times 10^{-5}$ & 4096 & ~ \\
         & 50 & 500 & 0.14 & 0.50 & 6.00 & $1.5 \times 10^{-5}$ & 4096 & $\star$ \\ \bottomrule
    \end{tabular}
    \label{tab:common_hparams}
\end{table}

Typically, if the dataset has a high value of $n$, we sampled a higher number of off-manifold augmentations ($N_{\text{off}}$) for the same number of on-manifold points ($N_{\text{on}}$). This is because the volume of an off-manifold band of width $max\_norm$ in $n$ dimensions, would roughly be of the order of  $max\_norm^{(n - m)}$. Although the volume increases exponentially in $(n-m)$, because of small $max\_norm$, we required only modest increases in the number of off-manifold augmentations, $N_{\text{off}}$. For instance, for Intertwined Swiss Roll with $m=1$, as we go from $n=2$ to $n=500$, we are able to obtain extremely low classification error rates with just a $20\times$ increase in $N_{\text{off}}$.

In almost all cases, we obtained low error rates from the first set of parameters that we chose. However, we have performed hyperparameter tuning for a few settings (marked by $\star$ in Table \ref{tab:common_hparams}) and found that it is possible to improve loss statistics further. We searched for the optimal values of $N_{\text{off}}$ ($\in [0.05, 6.00]\times 10^6$), learning rate ($\in [0.01, 8.00]\times 10^{-5}$), and batch size ($\in \{512, 2048, 4096\}$). Due to resource and time constraints, we could not tune hyperparameters for the other settings. Further hyperparameter tuning may improve performance in the other settings as well.

\section{More Results}

\subsection{Distance Prediction Accuracy}

\begin{figure}[t]
  \centering
  \begingroup
  \setlength{\tabcolsep}{4pt}
  \begin{tabular}{ccc}
    \includegraphics[width=0.31\linewidth]{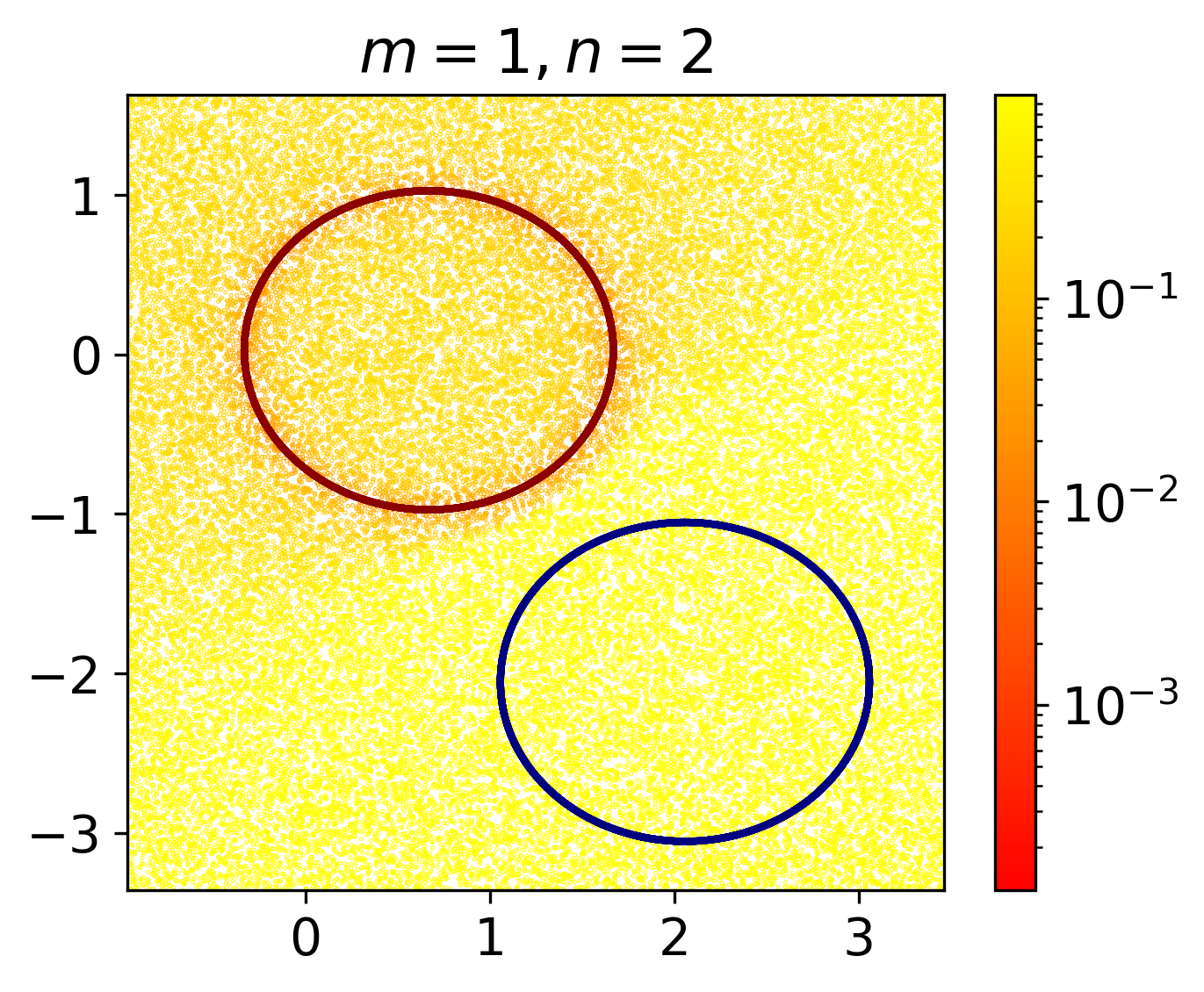} &
    \includegraphics[width=0.31\linewidth]{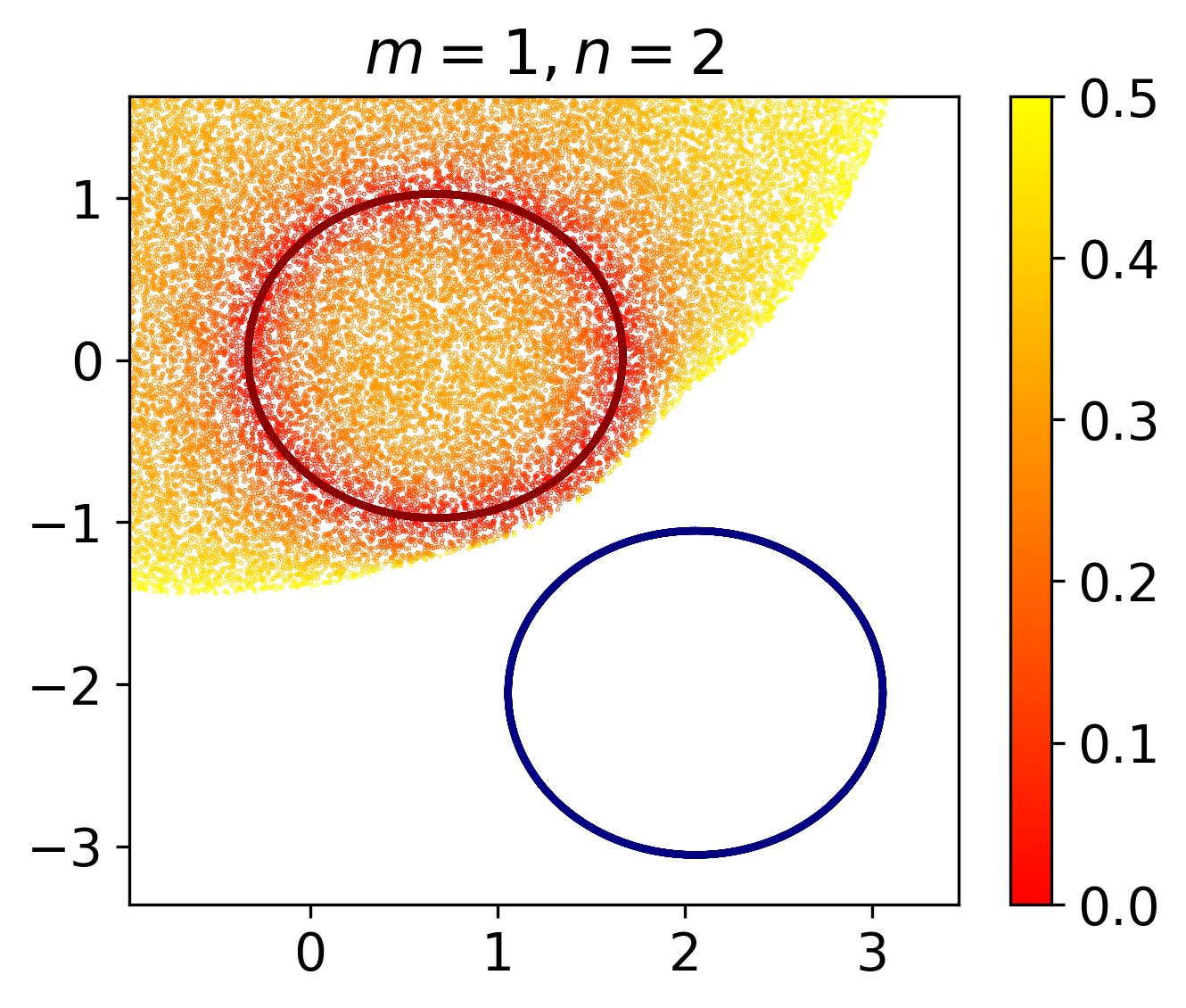} &
    \includegraphics[width=0.31\linewidth]{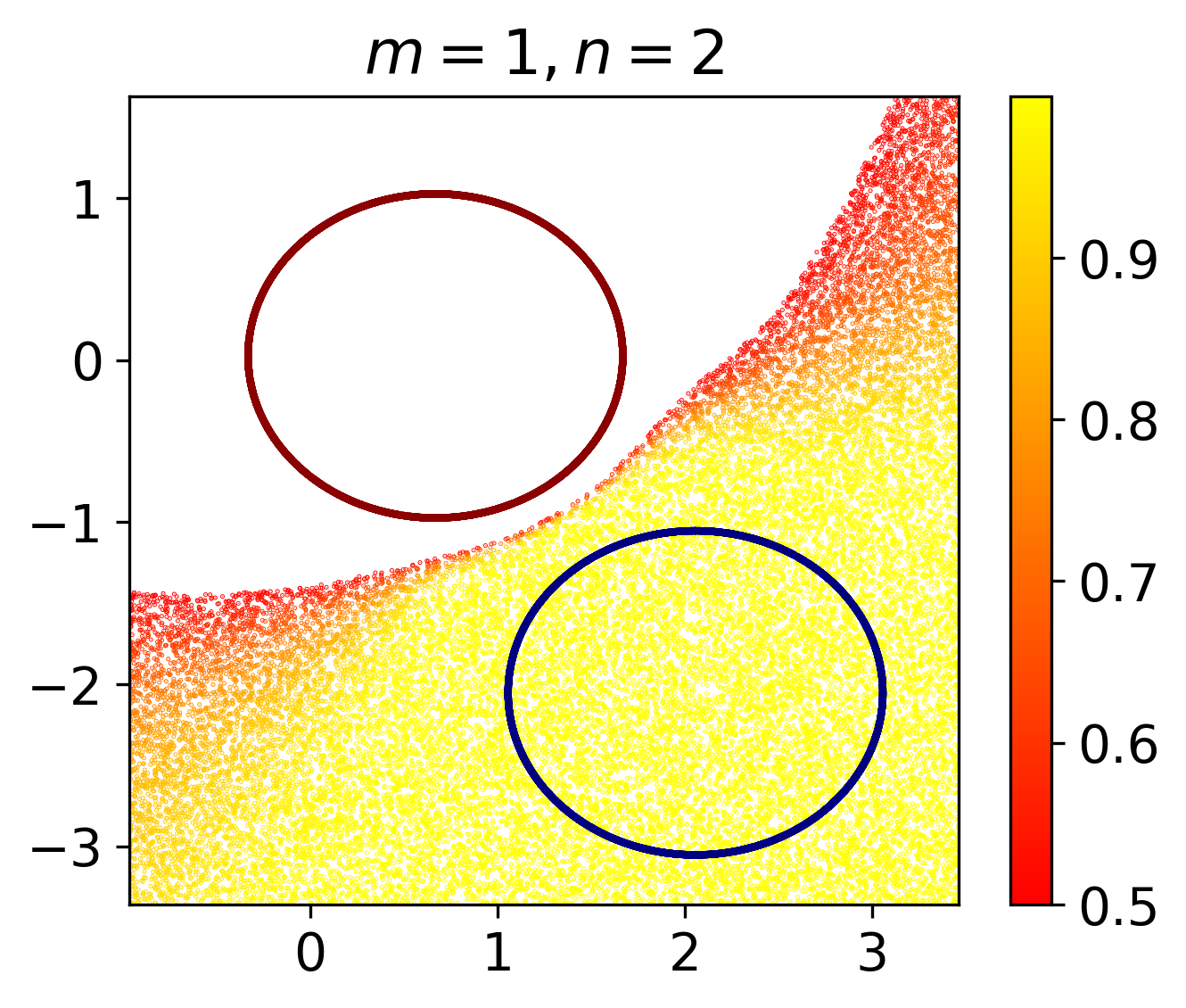}
    \\
    \includegraphics[width=0.31\linewidth]{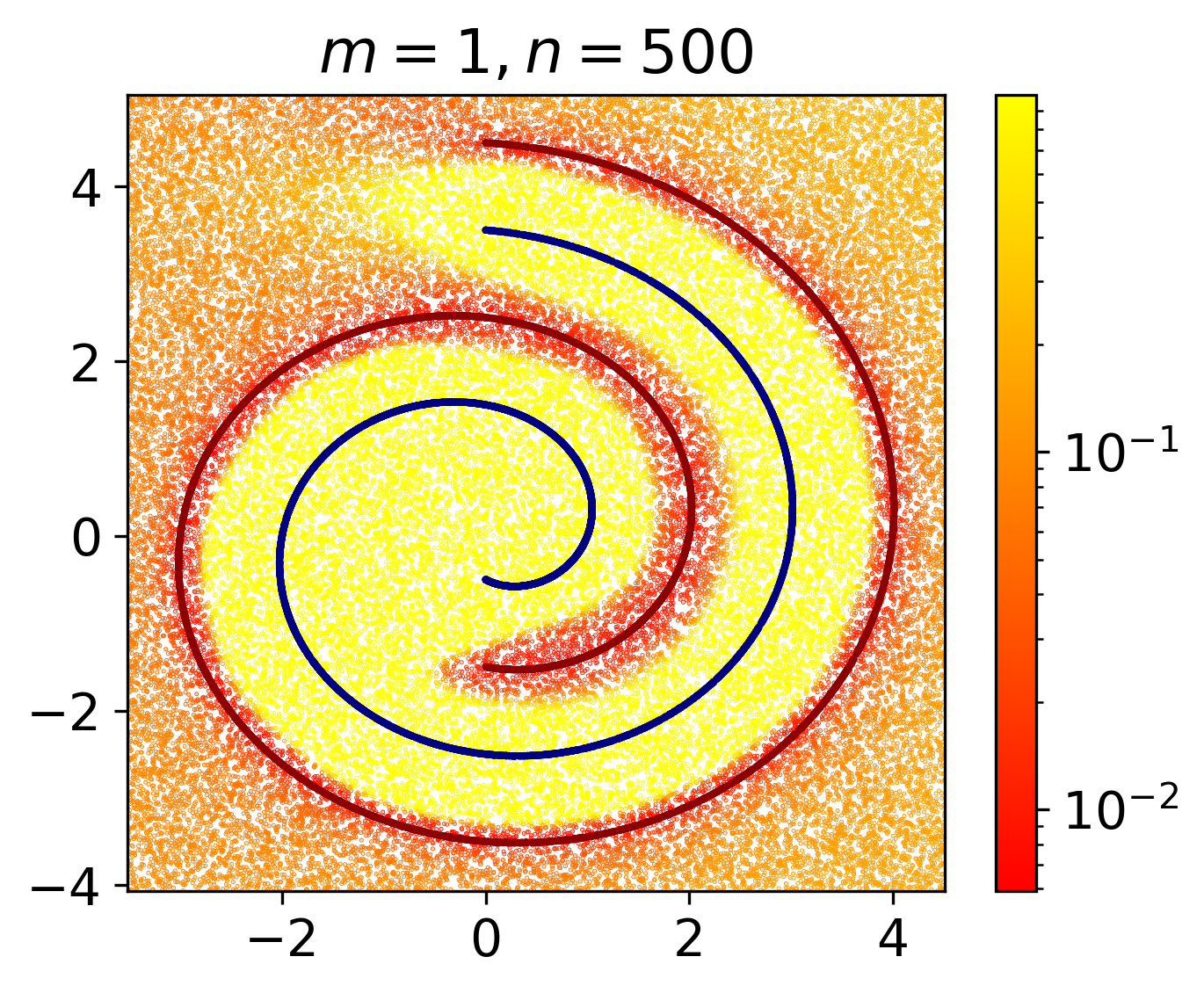} &
    \includegraphics[width=0.31\linewidth]{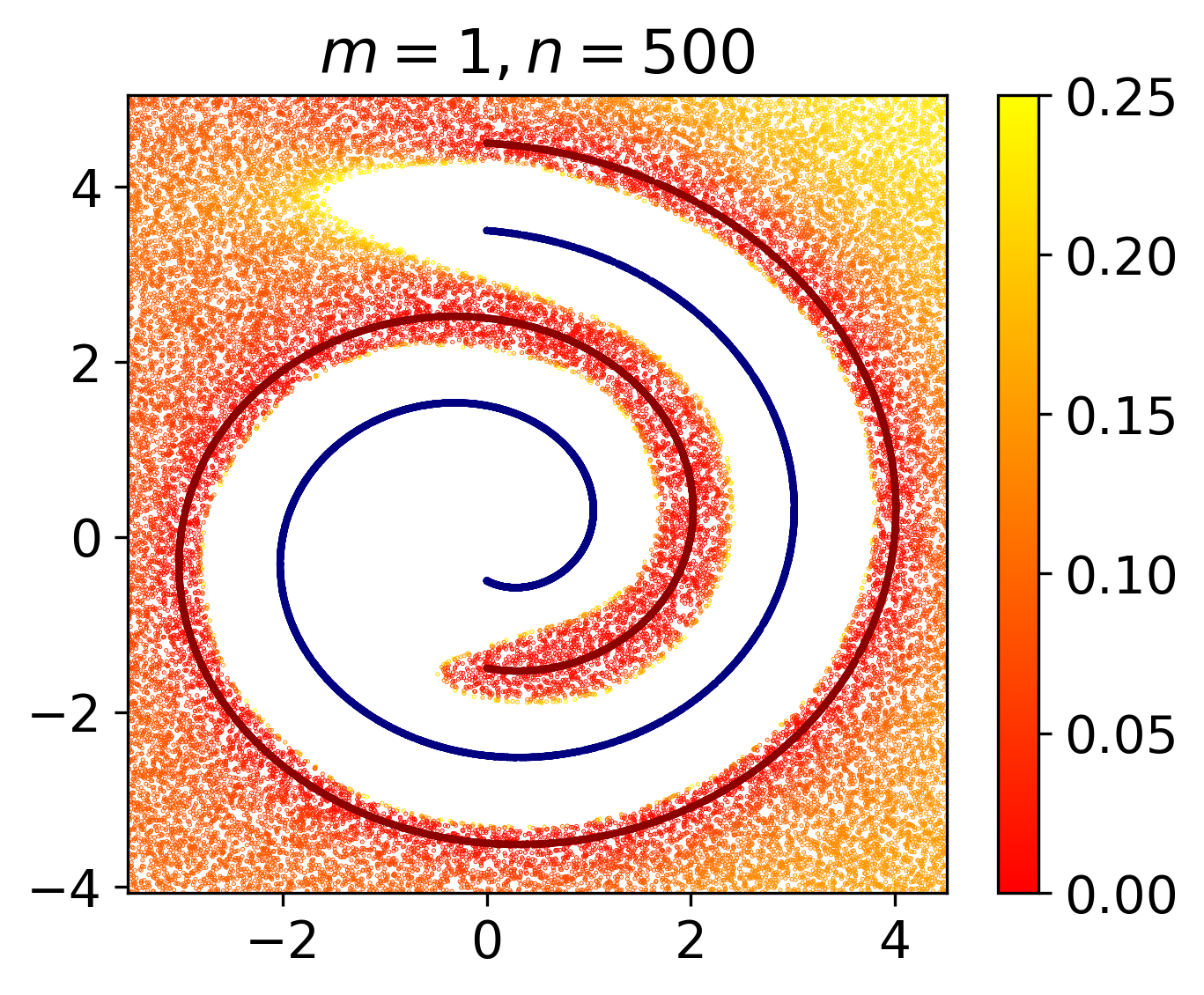} &
    \includegraphics[width=0.31\linewidth]{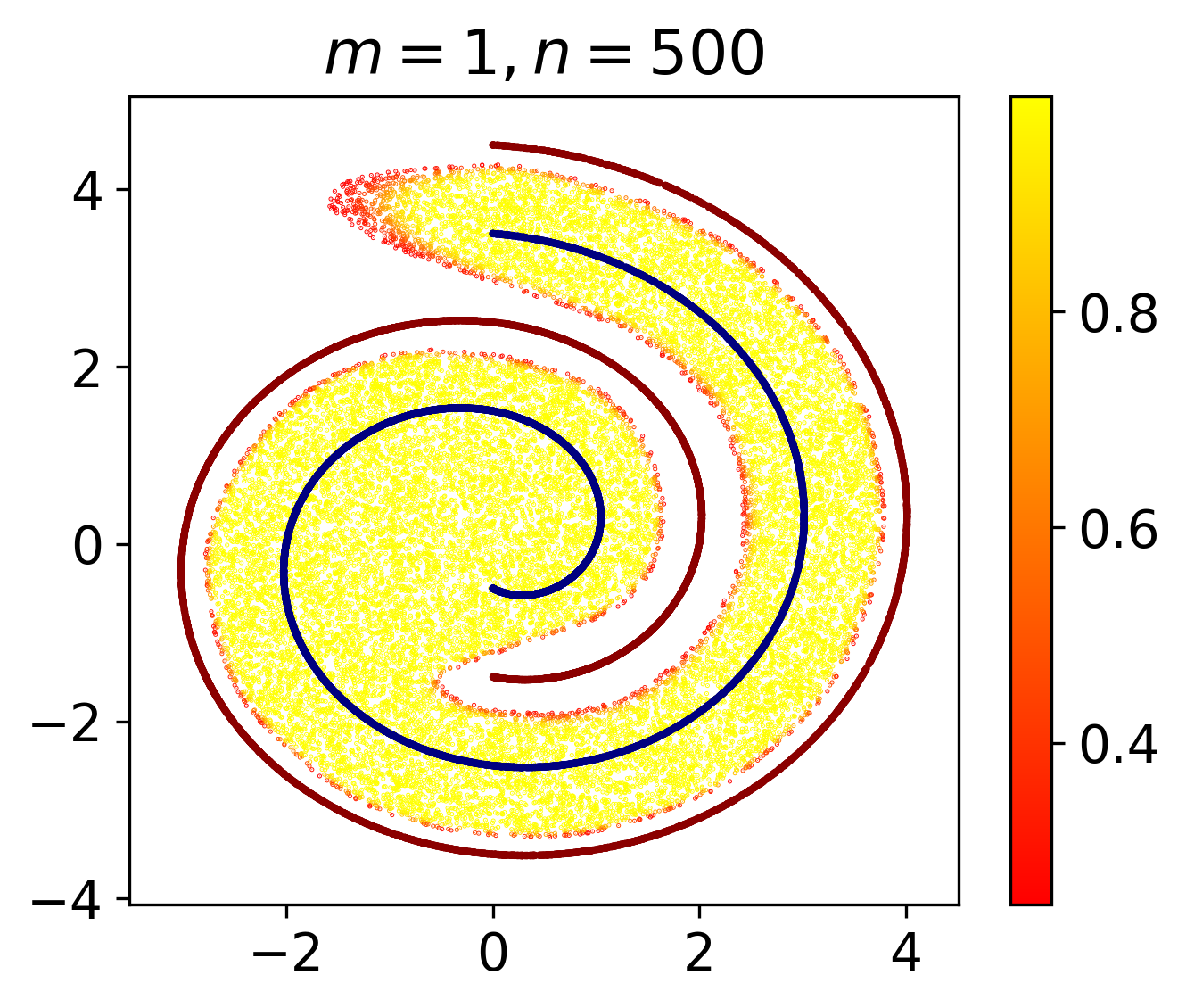}
    \\
    \includegraphics[width=0.31\linewidth]{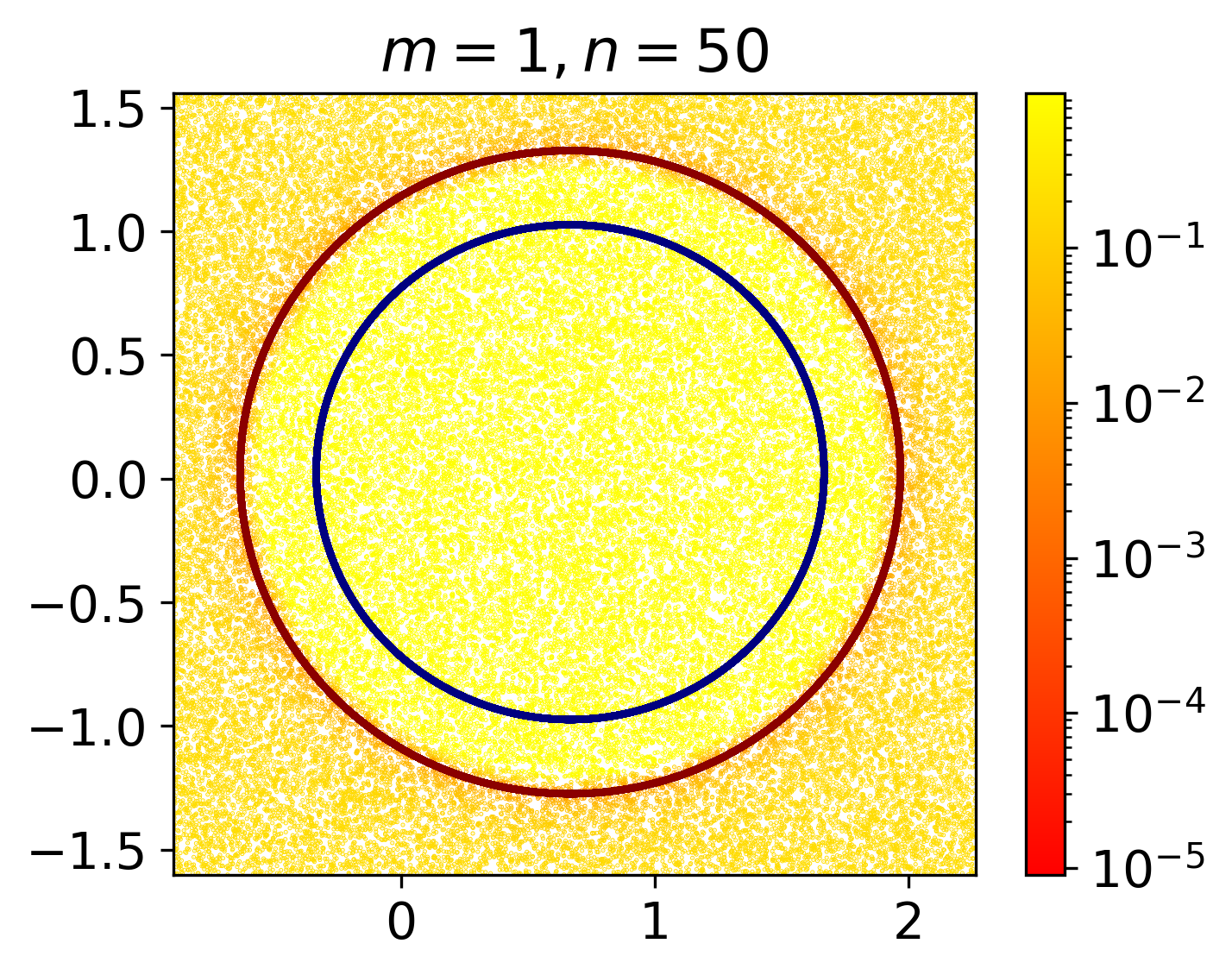} &
    \includegraphics[width=0.31\linewidth]{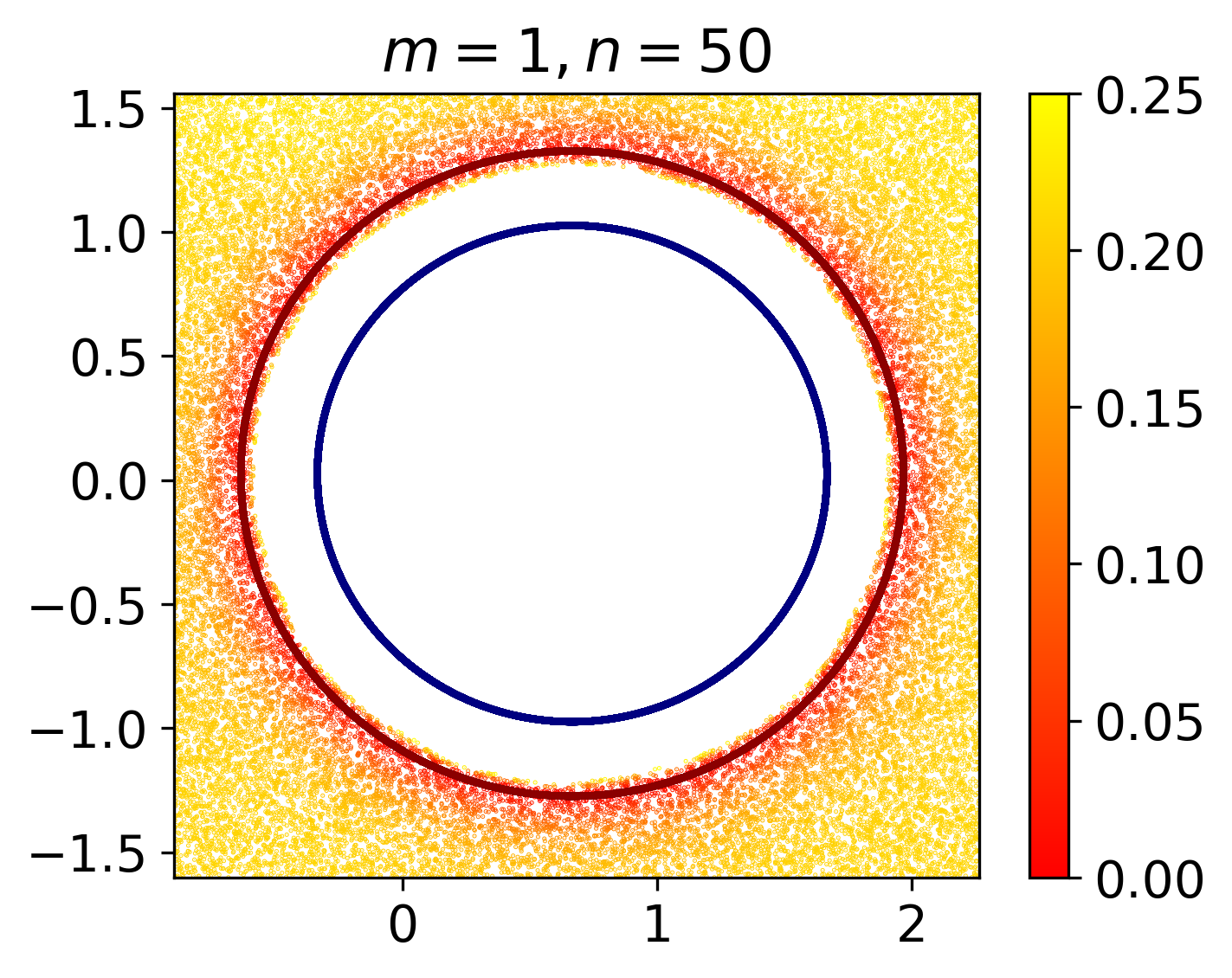} &
    \includegraphics[width=0.31\linewidth]{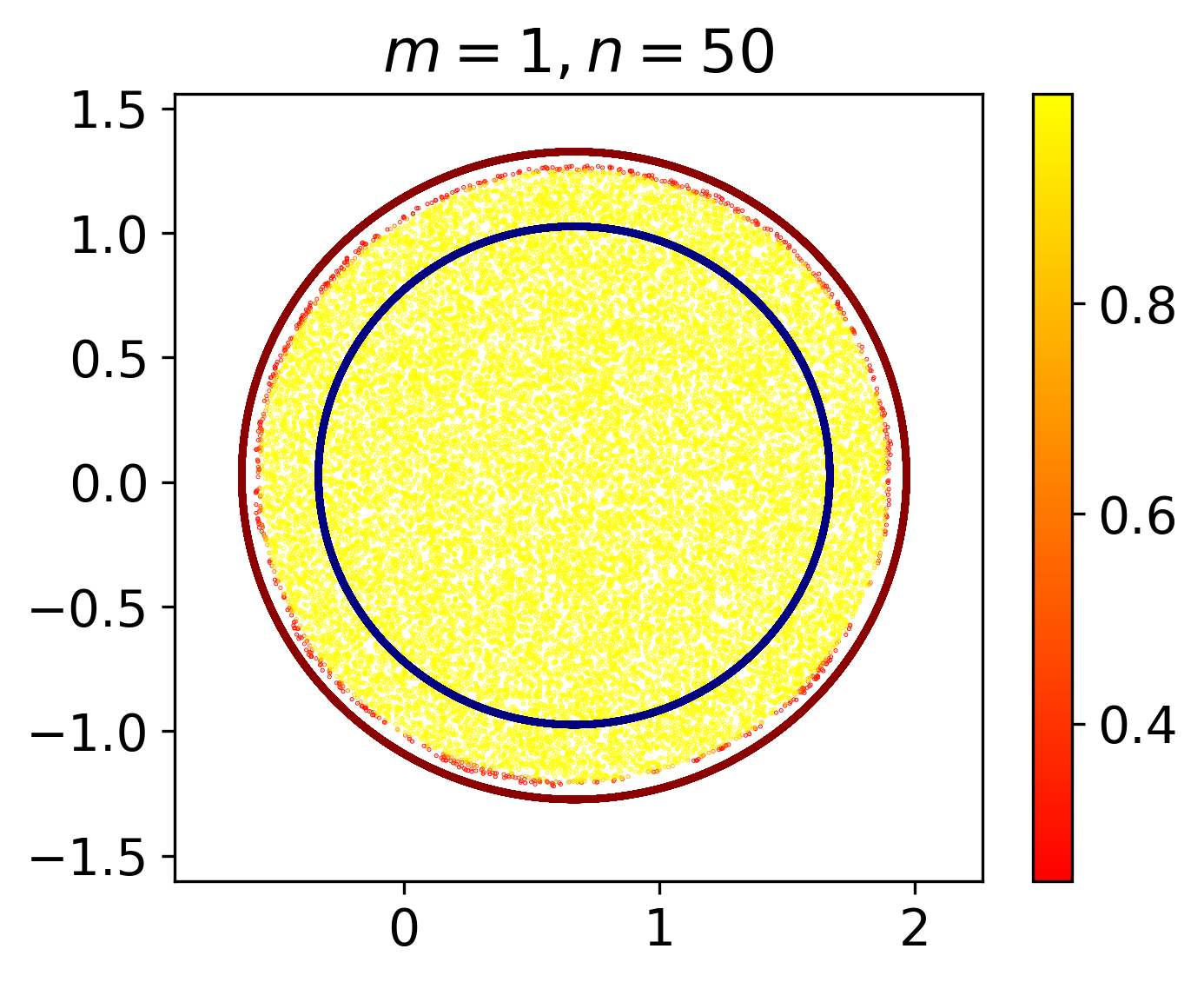}
    \\
    {(a) All points} & {(b) Lower Distance Points} & { (c) Higher Distance Points}
  \end{tabular}
  \endgroup
  \caption{Learnt Distances: Heatmap of predicted distance from the red class manifold. \textbf{(a)} Heatmap with all points, where the colors are decided by mapping the predicted values to colors on a log scale. \textbf{(b)} As with Figure \ref{fig:dist_heatmap}, we show the heatmap of points where the predicted value of distances are $\leq .5$ ($1^{\text{st}}$ row) and $\leq .25$ ($2^{\text{nd}} ~\&~3^{\text{rd}}$ rows). \textbf{(c)} Heatmap of points complimentary to those shown in (b), i.e. points with predicted distances $> .5$ ($1^{\text{st}}$ row) and $> .25$ ($2^{\text{nd}}~\&~3^{\text{rd}}$ rows).}
  \label{fig:distance_hmap_supple}
\end{figure}

\noindent\textbf{Heatmap of Learnt Distances} Figure \ref{fig:dist_heatmap}, showed heatmaps of predicted distances from the red manifold for points whose predicted distances were below a threshold. This was done because the \distlearner{} is trained to predict the distance of far-off points as $high\_distance$. As a result, the resultant heatmap of the entire region was such that on plotting all values, the smooth, increasing nature of the learnt distance function close to the manifold was not visible. 

In Figure \ref{fig:distance_hmap_supple}, we show the plots from Figure \ref{fig:dist_heatmap} again, along with the complete heatmap in the full region (log scale), as well as the complement of the point set plotted in Figure \ref{fig:dist_heatmap}. We can further see from Figure \ref{fig:distance_hmap_supple} that the \distlearner{} has indeed learned a smoothly increasing distance function, even in regions far away from the training data. For points closer to the other (blue) manifold, we can see that the \distlearner{} predicts a very high value (Figure \ref{fig:distance_hmap_supple}(c)). 

\subsection{Decision Regions}

\begin{figure}
\vspace{-6pt}
  \centering
  \begin{subfigure}[b]{0.5\linewidth}
    \includegraphics[width=\linewidth]{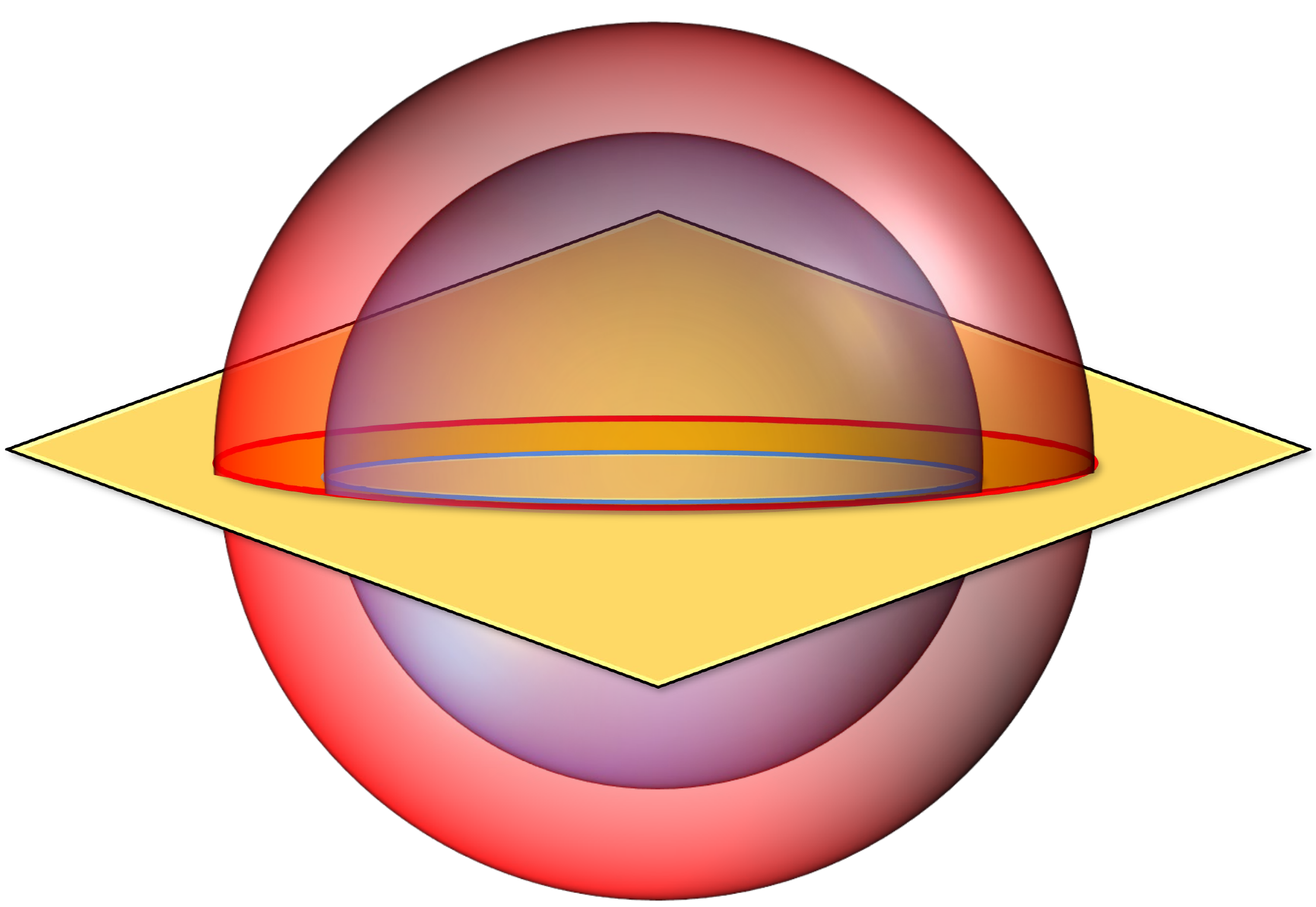}
    \caption{}
  \end{subfigure} \hspace{.15\linewidth}
  \begin{subfigure}[b]{0.25\linewidth}
    \includegraphics[width=\linewidth]{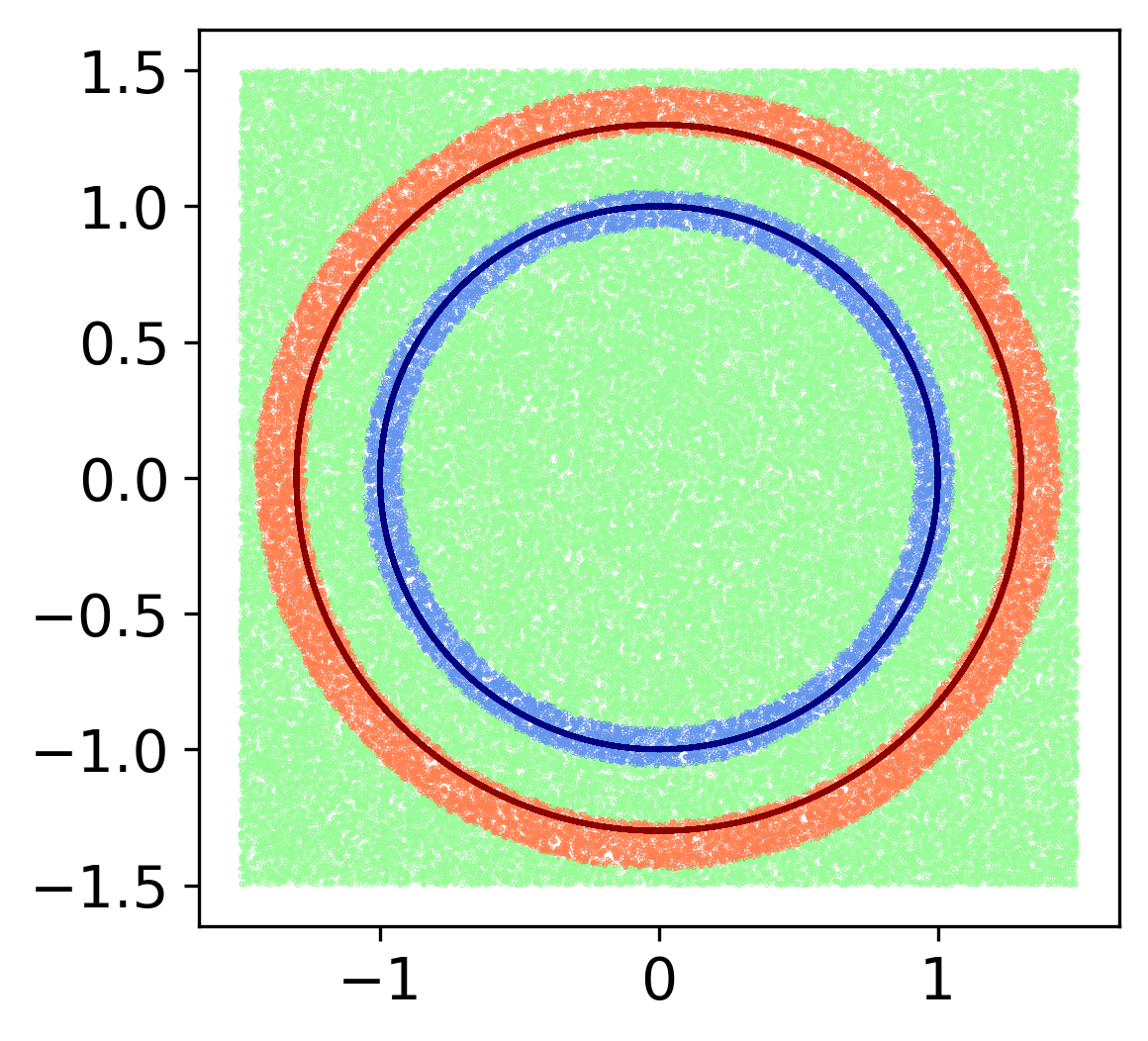} \\[-\lineskip]
    \includegraphics[width=\linewidth]{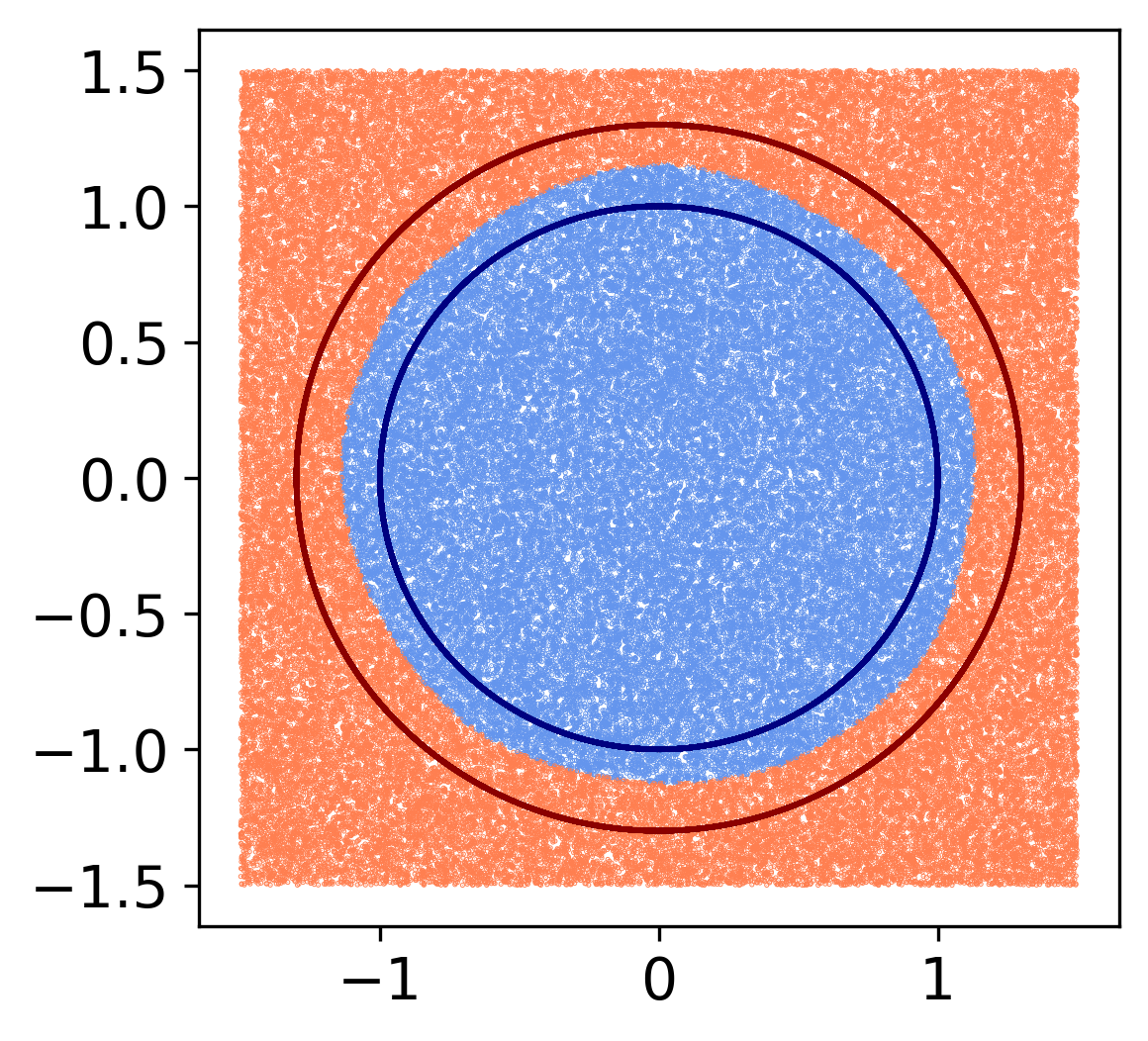}
    \caption{}
  \end{subfigure}

  \caption{\textbf{(a)} Illustration of a plane used for plotting decision regions for concentric spheres dataset with $m=2,~n=500$. \textbf{(b)} Decision regions for \distlearner{} (top) and \stdclf{} (bottom).}
  \label{fig:decreg_m2}
\end{figure}

In Figure \ref{fig:decision_boundary}, we visualized decision regions only for $m=1$ dimensional manifolds. Here we show decision regions for Concentric Spheres with $m=2$ (Figure~\ref{fig:decreg_m2}). We visualize the same by sampling points on a plane passing through the centre of the spheres and inferring them via \distlearner{} and \stdclf{}.

\subsection{Adversarial Robustness}
\label{sec:appenpdix_adversarial_robustness}
\begin{figure}[t]
    \centering
    \includegraphics[width=0.4\linewidth]{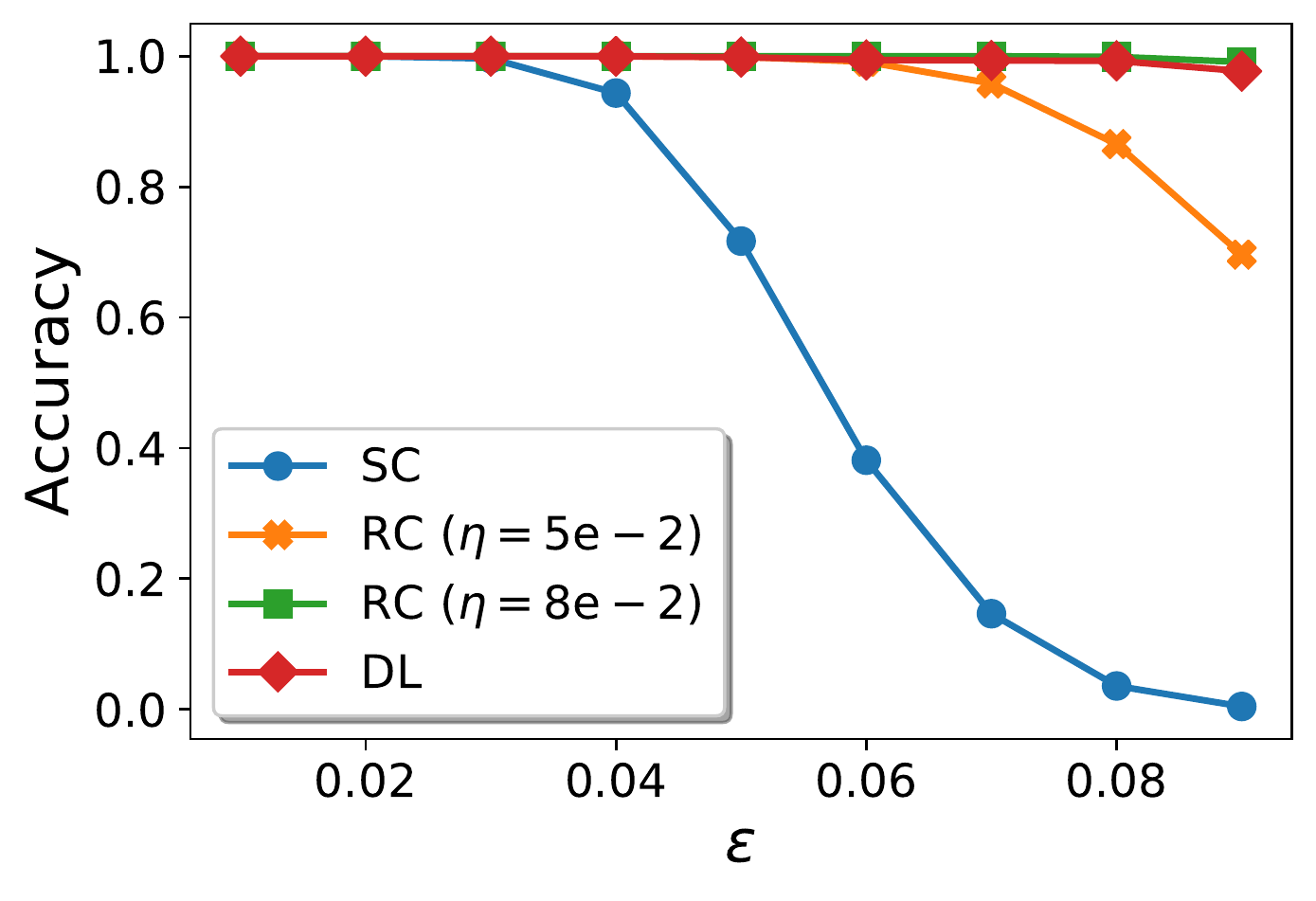}
    \caption{Adversarial Robustness of \distlearner{} (\textbf{DL}), \stdclf{} (\textbf{SC}), and \advclf{} (\textbf{RC}) over different values of $\epsilon$ (maximum adversarial perturbation) for $m=25,~n=500$.}
    \label{fig:adversarial_robustness2}
\end{figure}

We conducted our adversarial robustness experiments with another setting of the Concentric Spheres Dataset and chose $m=25,~n=500$ (all other settings were exactly the same as section \ref{sec:adversarial_robustness}). The results are shown in Figure~\ref{fig:adversarial_robustness2}. As shown, \distlearner{} continues to outperform the standard classifier significantly, and performss at par with adversarially trained classifiers.

\end{document}